\documentclass[10pt]{article} %
\usepackage[accepted]{tmlr}

\usepackage{amsmath,amsfonts,bm}

\newcommand{\newterm}[1]{{\bf #1}}

\def\Figref#1{Figure~\ref{#1}}

\def\eqref#1{equation~\ref{#1}}

\def\1{\bm{1}}

\def\mH{{\bm{H}}}

\def\mZ{{\bm{Z}}}

\DeclareMathAlphabet{\mathsfit}{\encodingdefault}{\sfdefault}{m}{sl}
\SetMathAlphabet{\mathsfit}{bold}{\encodingdefault}{\sfdefault}{bx}{n}

\newcommand{\softmax}{\mathrm{softmax}}

\newcommand{\modelname}{Feature Attending Recurrent Modules}
\newcommand{\modelabb}{FARM}

\newcommand{\ourmodel}{\modelabb}

\newcommand{\mb}[1]{\mathbb{#1}}

\newcommand{\numschema}{n}

\newcommand{\twoline}[2]{\begin{tabular}{@{}c@{}}#1 \\ #2\end{tabular}}

\newcommand{\cmark}{\ding{51}}%
\newcommand{\xmark}{\ding{55}}%
\newcommand{\Cmark}{{\color{green!50!black}\cmark}\xspace}%
\newcommand{\Xmark}{{\color{red!70!black}\xmark}\xspace}%

\definecolor{revision_color}{HTML}{000000}
\definecolor{revision_color2}{HTML}{000000}
\newcommand{\revision}[1]{\textcolor{revision_color2}{#1}}

\newcommand{\revisionthr}[1]{\textcolor{revision_color}{#1}}

\usepackage{hyperref}
\usepackage{url}
\usepackage{array}
\usepackage{booktabs}
\usepackage{graphicx}
\usepackage{caption}
\usepackage{enumitem} %
\usepackage{subcaption}
\usepackage{wrapfig}
\usepackage{lineno}
\usepackage[tracking=true]{microtype}
\usepackage{amssymb, pifont}
\usepackage{xspace} 
\usepackage{xcolor}
\usepackage{algorithm2e}

\title{Feature-Attending Recurrent Modules for Generalization in Reinforcement Learning}

\author{%
Wilka Carvalho\thanks{Contact author: \href{wcarvalho@g.harvard.edu}{wcarvalho@g.harvard.edu}. Work done during internship. Codebase: \href{https://github.com/wcarvalho/farm}{https://github.com/wcarvalho/farm}.}\,\,$^{,1}$ \hspace{1em}
Andrew Lampinen$^{2}$          \hspace{1em} \\
Kyriacos Nikiforou$^{2}$          \hspace{1em} 
Felix Hill$^{2}$          \hspace{1em}
Murray Shanahan$^2$          \hspace{1em}
\\[2pt]
$^1$Harvard University \hspace{2em}
$^2$Google DeepMind \hspace{2em}
}

\def\month{08}  %
\def\year{2023} %
\def\openreview{\url{https://openreview.net/forum?id=j4y3gN7VtW}} %

\begin{document}

\maketitle

\begin{abstract}
Many important tasks are defined in terms of objects.
To generalize across these tasks, a reinforcement learning (RL) agent needs to exploit the structure that the objects induce.
Prior work has either hard-coded object-centric features, used complex object-centric generative models, or updated state using local spatial features.
However, these approaches have had limited success in enabling general RL agents.
Motivated by this, we introduce ``Feature-Attending Recurrent Modules'' (FARM), an architecture for learning state representations that relies on simple, broadly applicable inductive biases for capturing spatial and temporal regularities.
FARM learns a state representation that is distributed across multiple modules that each attend to spatiotemporal features with an expressive feature attention mechanism.
We show that this improves an RL agent's ability to generalize across object-centric tasks.
We study task suites in both 2D and 3D environments and find that FARM better generalizes compared to competing architectures that leverage attention or multiple modules.

\end{abstract}

\begin{figure}[!htb]
  \centering
  \includegraphics[width=.975\textwidth]{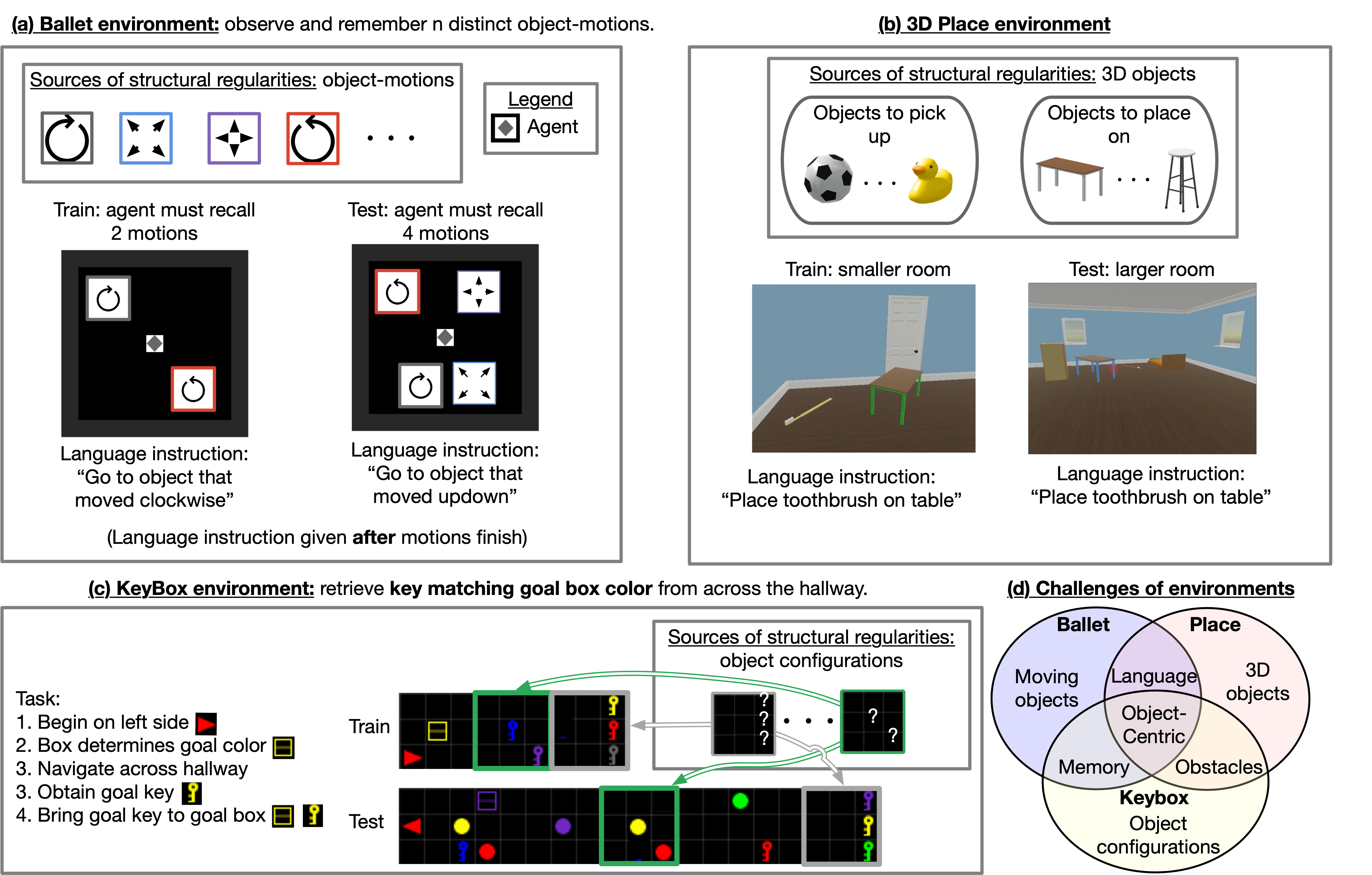}
  \vspace{-5pt}
  \caption{
    \textbf{\revisionthr{Three environments with different structural regularities induced by objects}}.
    In the Ballet environment, tasks share regularities induced by object motions; in the KeyBox environment, they share regularities induces by object configurations; and in the Place environment, tasks share regularities induces by 3D objects.
    The Ballet and KeyBox environments pose learning challenges for long-horizon memory and require generalizing to more objects.
    The KeyBox and Place environments pose learning challenges in obstacle navigation and requires generalizing to a larger map.
    Videos of our agent performing these tasks: \href{https://bit.ly/3kCkAqd}{https://bit.ly/3kCkAqd}.
    See \S \ref{sec:preliminaries} for a description of the problem setting.
    }\label{fig:environments}
\end{figure}
\section{Introduction}

Objects are key to real-world tasks. For example, a self-driving car needs to represent the movements of other cars, and a household robot needs to recognize and use kitchen items.
In order to generalize across tasks with objects, a reinforcement learning (RL) agent should capture and exploit object-induced structure present across the tasks.

One way to capture this structure is in an agent's state representation.
Unfortunately, flexibly capturing objects in a state representation is challenging because an \revisionthr{objects have many dimensions that can vary}. 
Consider a household robot tasked with cooking. Completing the task might require {memory} about objects that range in size, shape, and color (e.g. a stove vs. a tomato). Additionally, objects in motion might require that the agent represent temporal information about the objects.
It is unclear how to best incorporate objects into a state representations to enable generalization.

Prior work has attempted to capture object-induced task structure by hand-designing object-centric state features~\citep{diuk2008object, carvalho2020reinforcement,  borsa2018universal, Marom2018ZeroShotTW}.
The ``COBRA'' agent~\citep{watters2019cobra} avoids hand-designing features by learning an object-centric generative model. 
However, these methods are limited in their generality because they rely on relatively strong inductive biases. For example, COBRA relies on environments being fully-observable and objects having regular shapes to learn representations by predicting object segmentations.
We focus on weak inductive biases in order to maximize an architecture's flexibility.

Objects can be described by subsets of features over space and time.
We conjecture that weak inductive biases for capturing subsets of features over space and time may enable agents that can flexibly incorporate objects into state across a wide range of environments.

We propose \textit{\modelname~(\modelabb)}, a simple but flexible architecture for learning state representations when tasks share object-induced structural regularities. \modelabb~learns state representations that are \emph{distributed} across multiple, smaller recurrent modules.
To help motivate this, consider word embeddings.
A word embedding can represent more information than a one-hot encoding of the same dimension because subsets of dimensions can coordinate activity to represent different patterns of word usage.
Analogously, learning multiple modules enables \modelabb~to coordinate subsets of modules to represent different temporal segments in an agent's experiences. 
To capture general object-induced patterns, modules select observation information to update with by \revisionthr{applying a mask to the channels of spatiotemporal observation features}.

We study \modelabb~across three diverse object-centric environments, each with their own suite of tasks that share object-induced structural regularities.
Tasks in the Ballet environment share regularities induced by object motions; tasks in the Place environment share regularities induced by navigating towards  and around 3D objects; and tasks in the KeyBox environment share regularities induced by object configurations.
These environments present a number of challenges.
First, their state-space grows exponentially with the number of objects.
The more distractor objects an environment has, the larger the chance an object will obstruct an agent's path. This requires learning a policy that can navigate around distractor-based obstacles.
When task objects appear in sequence, this can require long-horizon memory of object information (e.g. of goal information).
Finally, tasks defined by language can require an agent learn a complex mapping (e.g. to object motions and to irregular shapes in our tasks).
Across these environments, we study an agent's ability to recombine object-oriented memory, obstacle-avoidance, and navigation to longer tasks with more objects.

We compare against methods with weak inductive biases for enabling objects to emerge in a state representation.
Recent work has shown that spatial attention is a simple inductive bias for strong performance on object-centric vision tasks because it enables attending to individual objects~\citep{greff2020binding, locatello2020object, goyal2020object}.
Thus, we compare against recent RL agents that leverage spatial attention for object-centric state-update functions~\citep{goyal2019recurrent, mott2019towards}.

\textbf{Our core contribution is} to show that we can improves an RL agent's ability to generalize to out-of-distribution tasks by having multiple modules attend to spatiotemporal features with feature attention. We expand on this below:
\begin{enumerate}[labelindent=0cm,labelsep=2pt,noitemsep,nolistsep, topsep=0pt,leftmargin=*]
\setlength{\itemsep}{1pt}
\setlength{\parskip}{0pt}
\item FARM leverages multiple modules that each apply feature-wise attention to spatiotemporal features. This enables generalizing (a) memory to longer combinations of object motions (\S \ref{sec:experiments-ballet}); (b) navigation to 3D objects in larger environments (\S \ref{sec:experiments-playroom}); and (c) memory of goal information to longer tasks with more distractors (\S \ref{sec:experiments-keybox}).
\item Competing methods have modules which leverage spatial attention, which has been shown to enable object-centric state updates. Across diverse object-centric RL tasks, we find that spatial attention has mixed benefits and can interfere with the benefits of learning multiple modules.
\item We hypothesize that FARM enables an RL agent to generalize to combinations of its experience by representing different temporal segments across subsets of modules (see \Figref{fig:architecture-intro}). In \S\ref{sec:analysis}, we analyze \modelabb~and provide evidence that object-induced temporal regularities are indeed represented across subsets of modules.
\end{enumerate}

\section{Related work on generalization in deep RL}\label{sec:background}
The key question for generalization is how to capture structure in the problem in a flexible way. How much structure do you build in? How much do you let the agent discover? Some work takes a data-driven approach~\citep{tobin2017domain, packer2018assessing, hill2020environmental, justesen2018illuminating}. Others have a policy that captures task structure with either hierarchical RL~\citep{Oh2017ZeroShotTG,zhang2018composable,sohn2018hierarchical,sohn2020meta,brooks2021reinforcement} or successor features~\citep{borsa2018universal, barreto2020fast}. A final strand focuses on learning invariant representations~\citep{higgins2017darla,chaplot2018gated,Lee2020NetworkRA,zhang2021learning} or building in inductive biases~\citep{mott2019towards, goyal2019recurrent}.
In this work we focus on weak inductive biases for capturing structure. Below we review approaches most closely related to ours.

\textbf{Generalizing across object-centric tasks} dates back at least to object-oriented MDPs~\citep{dvzeroski2001relational,diuk2008object} which enabled generalization by representing dynamics with logical object attributes~\citep{kansky2017schema, Marom2018ZeroShotTW}. Successor features have also leveraged objects for generalization by formulating rewards as linear with object-centric features~\citep{borsa2018universal, barreto2020fast}. A common thread among these directions is that they relied on hand-designed object features. 
\citet{watters2019cobra} avoided hand-designing features by learning an object-centric generative model~\citep{burgess2019monet}. However, they focused on fully-observable top-down environments with regular shapes, which allowed them to predict future object masks.
This is incompatible with our  environments. 
While research on object-centric models~\citep{kabra2021simone, zoran2021parts} has progressed, these methods commonly add training complexity (more objective terms, extra modules, etc.) and make stronger assumptions (e.g. on the number of objects).
We differ from this work because we focus on simple, broadly applicable inductive biases for capturing object-induced task regularities.

\textbf{Generalizing with feature attention} has also been studied by~\citet{ chaplot2018gated}. 
\revisionthr{They showed that mapping language instructions to masks over the channels of observation features enabled generalization to language instructions with new feature combinations. While \modelabb~also learns a mask over observation features, our work has two important differences. First, we develop a multi-head version where different recurrent modules produce their own masks.} This enables \modelabb~to leverage this form of attention in settings where language instructions don't indicate what to attend to (this is true in $2/3$ of our tasks). Second, we are the first to show that feature attention enables generalizing memory of object motions and of goal information to longer tasks (\Figref{fig:performance-ballet} and \Figref{fig:performance-keybox}, respectively).

\textbf{Generalizing with top-down spatial attention}.
Most similar to \modelabb~are the Attention Augmented Agent (AAA)~\citep{mott2019towards} and Recurrent Independent Mechanisms (RIMs)~\citep{goyal2019recurrent}. 
Both are recurrent architectures that leverage spatial attention to learn an object-centric state-update function.
Both showed generalization to novel distractors.
{The major difference between AAA, RIMs, and \modelabb~is that \modelabb~attends to an observation with feature attention as opposed to spatial attention.}
Our experiments indicate that spatial attention has limited utility in updating state during reinforcement learning of tasks defined by object motions (\Figref{fig:performance-ballet}) or 3D objects (\Figref{fig:performance-playroom}). In terms of modularity, we also show different results from RIMs who showed that their modules ``specialize''. Our experiments suggest that in FARM, a modular state instead leads subset of modules to \emph{jointly} represent regularities in an agent's experience (\S\ref{sec:analysis}). 

\section{Problem setting}\label{sec:preliminaries}
We study generalization across tasks within deterministic, partially-observable, pixel-based environments. Within an environment, a task is defined by a Partially Observable Markov decision processes (POMDP): $\mathcal{M} = \langle 
\mathcal{S}, \mathcal{A}, \mathcal{O},
R, T, \psi
\rangle$.
$\mathcal{S}$ corresponds to environment states,
$\mathcal{A}$ corresponds to actions that agent can take,
$\mathcal{O}$ corresponds to the agent's observations,
$r = R(s,a) \in \mathbb{R}$ is the reward function,
$s'=T(s,a) \in \mathcal{S}$ is the environment transition function,
and
$o=\psi(s) \in \mathcal{O}$ is an observation function that maps the underlying environment state to an RGB image.

We seek an RL agent that learns to perform tasks by finding a policy $\pi$ that maximizes the expected discounted sum of rewards it obtains starting at a state $s$: $V(s) = \mathbb{E}\left[\sum^{\infty}_{t=0}\gamma^t R(S_t, A_t)\right]$---also known as the \emph{value} of a state. In a POMDP, the agent doesn't have access to the environment state. A common strategy is to instead learn an \emph{``agent state''} representation, $s_t^A$, that compresses the full history $(o_1, a_1, r_1, \ldots, a_{t-1}, o_t)$ into a sufficient statistic suitable for selecting actions. The {agent state} is commonly learned with a recursive function $s_{t}^A=\eta(o_t, a_{t-1}, r_{t-1}, s_{t-1}^A)$. 

\textbf{Object-induced structural regularities}. We study object-centric environments, where objects induce structural regularities across tasks in the reward functions $R$, transition functions $T$, and observation functions $\psi$. 
For example, consider the KeyBox environment in Figure~\ref{fig:environments} (c). 
First, $R(s,a)$ always specifies the goal key based on a goal box. 
Second, whenever the agent has to navigate around an obstacle (see \Figref{fig:architecture-intro}, b), the agent always sees the sprite it controls move closer to an object and then around it. 
This is true regardless of \emph{where} in the hallway the agent observes the obstacle because of regularities in the transition function $T(s,a)$ and observation function $\psi(s)$.
We want an agent that captures these regularities in its representation for state to enable zero-shot generalization to new tasks.

\section{Architecture:~\modelabb}\label{sec:architecture}
\begin{figure}[bht]
\centering
\includegraphics[width=.58\textwidth]{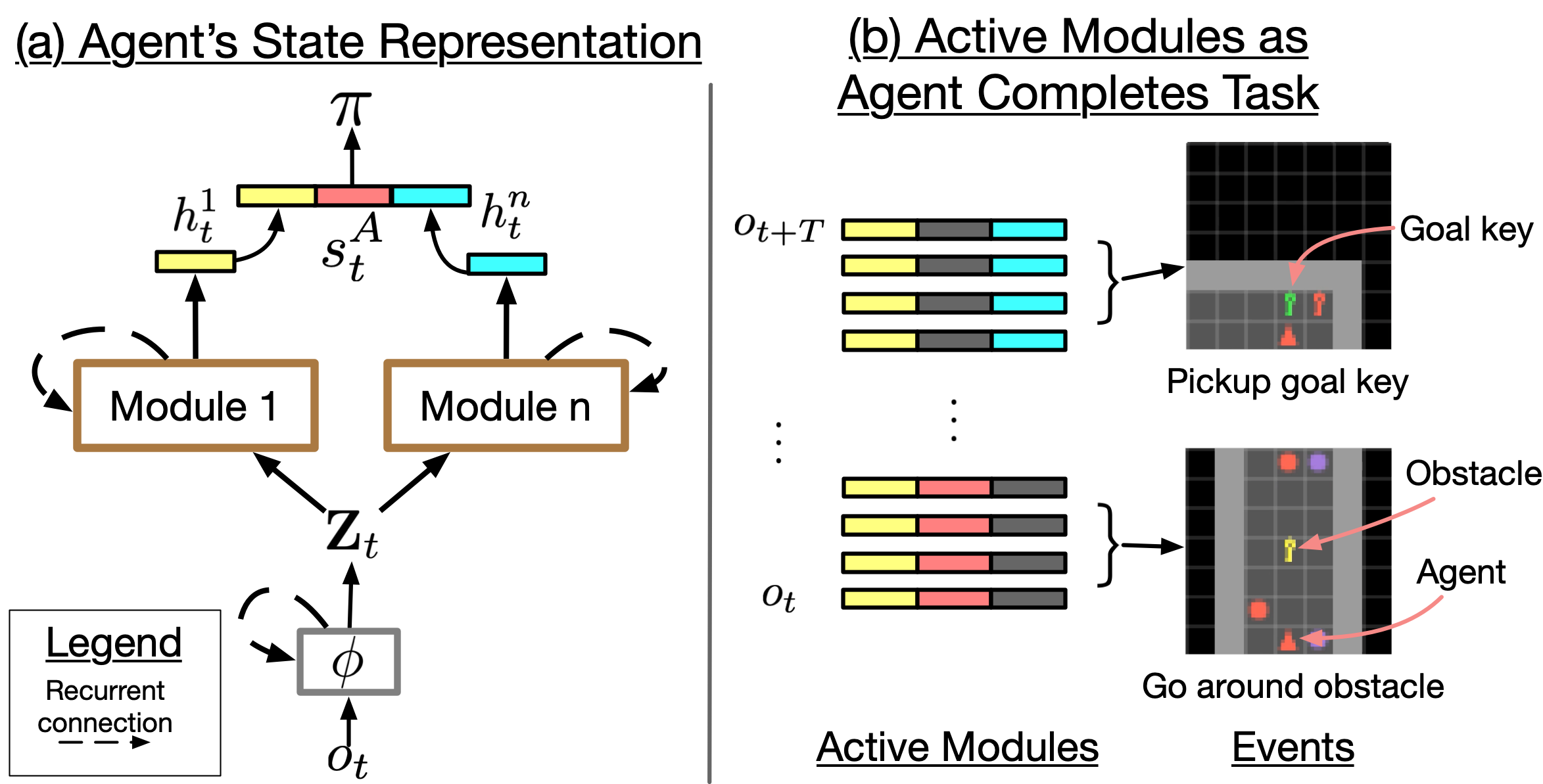}
\caption{
  \textbf{Overview of~\modelabb}.
  (a) \modelabb~learns an agent state representation that is distributed across $n$ recurrent modules.
  (b) By distributing agent state across multiple modules, \modelabb~is able to represent different object-centric task regularities, such as navigating around obstacles or picking up goal keys, across subsets of modules. We hypothesize that this enables a deep RL agent to flexibly recombine its experience for generalization. See \S\ref{sec:architecture} for details on the architecture and \S\ref{sec:analysis} for supporting analysis.
  }\label{fig:architecture-intro}
\end{figure}

We propose a new architecture, ``\modelname'' (\modelabb) for learning an agent's state representations when an environment has object-induced structural regularities. 
We provide an overview of the architecture in \Figref{fig:architecture-intro}. 
Instead of representing agent state with a single recurrent function, \modelabb~learns a state representation that is distributed across $n$ recurrent functions $\{\eta^{k}\}_{k=1}^n$, which we call modules (\Figref{fig:architecture-intro}, a). Distributing state across modules allows subsets of modules to jointly represent different regularities in the agent's experience (\Figref{fig:architecture-intro}, b). 
We hypothesize that having subsets of modules represent different regularities in the agent's experience enables the agent to flexibly recombine its experience for generalization.

At each time-step $t$, each module updates with both observation features and information from other modules.
First, the agent computes observation features with a recurrent observation encoder, $\mZ_{t} = \phi(o_t, \mZ_{t-1})$. 
Afterward, each module creates a \emph{query} vector by combining its previous module-state with the previous action and reward, $q^k_{t-1} = [h_{t-1}^{k}, a_{t-1}, r_{t-1}]$.
The query is used to attend to observation features via a dynamic feature attention mechanism $u^k_t = f^k_{\tt att}(\mZ_t, q^k_{t-1})$.
The query is also used to retrieve information from other modules with a transformer-style attention mechanism $\nu^k_t = f^k_{\tt share}(s^A_{t-1}, q^k_{t-1})$.
(We explain both attention mechanisms in more detail below).
Each module updates with both attention outputs to produce the next module-state $h_t^{k} = \eta^{k}(u^k_t, \nu^k_t, q^k_{t-1})$.
\revisionthr{If a task additionally has a language description $o_{\tt lang}$ (as 2 of our experiments do), the module update also updates with an embedding of this description, $z_{\tt lang} = f_{\tt lang}(o_{\tt lang})$}.
Agent state is then defined by the combination of these module-states $s_t^A = [ h_t^{1}, \ldots, h_t^{\numschema}]$.
We illustrate this in \Figref{fig:architecture-appendix} and summarize the computations below:
\begin{align}
  \mZ_t &= \phi(o_t, \mZ_{t-1}) && \text{obs features} \\
  q^k_{t-1} &=[h_{t-1}^{k}, a_{t-1}, r_{t-1}] && \text{query} \\
  u^k_t &= f^k_{\tt att}(\mZ_t, q^k_{t-1}) && \text{obs attention} \\
  \nu^k_t &= f^k_{\tt share}(s^A_{t-1}, q^k_{t-1}) && \text{share info} \\
  h_t^{k} &= \eta^{k}(u^k_t, \nu^k_t, q^k_{t-1}) && \text{module update} \\
  s_t^A &= [ h_t^{1}, \ldots, h_t^{\numschema}] && \text{agent state} 
\end{align}
\revisionthr{where $[\cdot]$ is an operation that concatenates input vectors into a long vector.}
We now describe each computation in more detail.

\begin{figure*}[!htb]
\centering
\includegraphics[width=.9\textwidth]{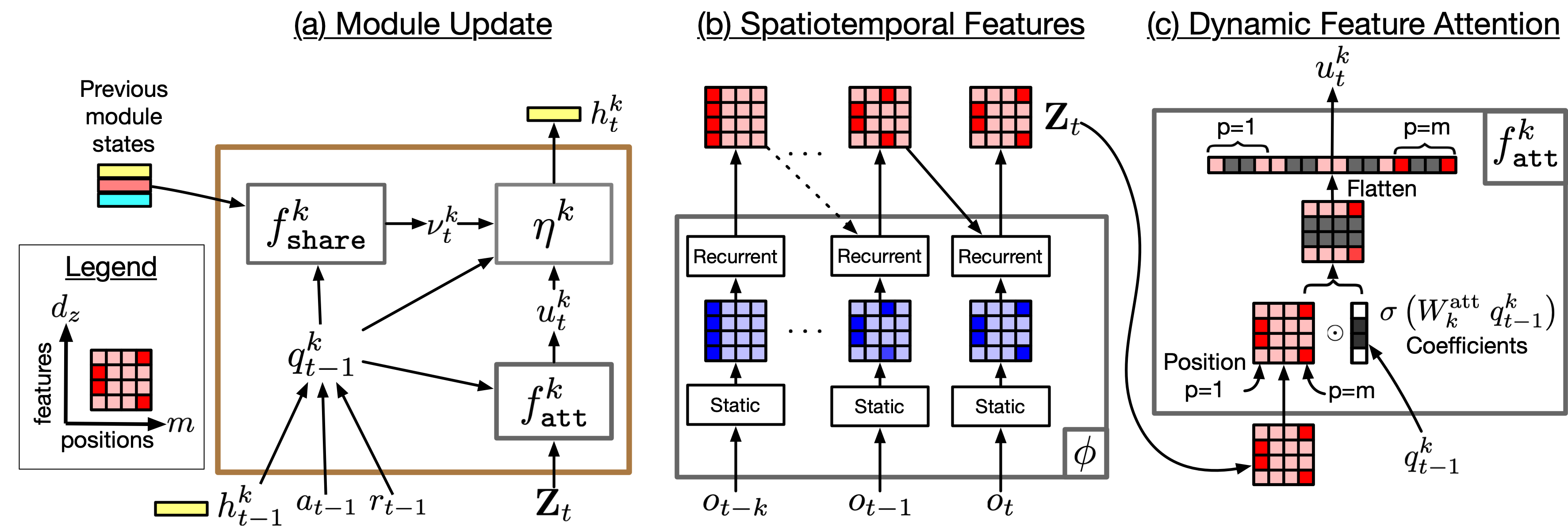}
\caption{
  \textbf{Computations of~\modelabb}.
  (a) Schematic of updates. See 2nd paragraph in \S\ref{sec:architecture} for details.
  (b) In order to update with features that describe both visual and temporal regularities, the agent learns structured spatiotemporal features $\mZ_t \in \mb{R}^{m \times d_z}$ that share $d_z$ spatio-temporal features across $m$ spatial positions. Here we show toy computations where static observations features (blue) on the top and bottom row move to spatial positions to the right. The resultant spatio-temporal features (red) also include temporal information about the features (here, that the features came from leftward spatial positions).
  (c) $f^k_{\tt att}$ computes coefficients for features and applies them uniformly across all spatial positions. This allows the agent to attend to all spatial position that possess desired features.
  }\label{fig:architecture-update}
\end{figure*}

\textbf{Structured spatiotemporal observation features.}
Our first insight is that modules can attend to features describing an object's motion if an agent learns observation features that describe both spatial and temporal regularities.
An agent can accomplish this by learning structured spatiotemporal features with a recurrent observation encoder $\mZ_t = \phi(x_t, \mZ_{t-1})\in \mathbb{R}^{m \times d_z}$ that share $d_z $ features across $m$ spatial positions\footnote{One can convert height by width observation features as follows: $\mathbb{R}^{h \times w \times d_z} \rightarrow \mathbb{R}^{hw \times d_z}$}. At each spatial position, these features both describe what is there visually along with temporal information about the recent dynamics of these features.
We show example toy computations in \Figref{fig:architecture-update} (b).

\textbf{Dynamic feature attention.} 
Our second insight is that feature attention is an expressive attention function that can focus on desired information present across all spatial positions in observation features. An agent accomplishes this by having a module predict feature coefficients that it applies to uniformly across all spatial positions in $\mZ_t$~\citep{perez2018film,chaplot2018gated}. We show example toy computations in \Figref{fig:architecture-update} (c). We found it useful to linearly project the features before and after using shared parameters as in~\citet{andreas2016neural, hu2018squeeze}. The operations are summarized below:
\begin{align}\label{eq:attn}
  f^{k}_{\tt att}(\mZ_t, q^k_{t-1}) = \left(\mZ_t W_{1} \odot 
  \sigma(W^{\tt att}_{k} q^k_{t-1})
    \right) W_{2}
\end{align}
where $\odot$ denotes an element-wise product over the feature dimension and $\sigma$ is a sigmoid non-linearity.
Since our features capture dynamics information, this allows a module to attend to object motion (\S\ref{sec:experiments-ballet}).
When updating, we flatten the attention output. Flattening leads all spatial positions to be treated uniquely and allows a module to represent aspects of the observation that span multiple positions, such as 3D objects (\S\ref{sec:experiments-playroom}) and spatial arrangements of objects (\S\ref{sec:experiments-keybox}).
Since the feature-coefficients for the next time-step are produced with observation features from the current time-step, modules can \emph{dynamically shift} their attention when task-relevant events occur (see \Figref{fig:keybox-activity}, b for an example).

\textbf{Sharing information.} Similar to RIMs~\citep{goyal2019recurrent}, before updating, each module retrieves information from other modules using transformer-style attention~\citep{vaswani2017attention}. We illustrate this in \Figref{fig:architecture-appendix} (c).
We define the collection of previous module-states as $\mH_{t-1} = \left[h_{t-1}^{(1)}; \ldots; h_{t-1}^{(n)}; \mathbf{0} \right] \in \mb{R}^{(n+1) \times d_h}$, where $\mathbf{0}$ is a null-vector used to retrieve no information. 
A module computes a ``retrieval query'' to search for information as $q_r^k= W^{\tt que}_{k} q^k_{t-1} \in R^{d_h}$.
That module computes ``retrieval keys and values'' as $K^k=\mH_{t-1} W^{\tt key}_{k} \in R^{n+1 \times d_h}$ and $V^k=\mH_{t-1} W^{\tt val}_{k} \in R^{n+1 \times d_h}$, respectively.
Each module then retrieves information as follows:
\begin{align}\label{eq:share}
  f^k_{\tt share}(s^A_{t-1}, q^k_{t-1}) = \softmax \left(
    \frac{
      q_r^k {K^k}^{\top}}{\sqrt{d_h}}\right) V^k.
\end{align}
Intuitively, the dot-product inside the softmax is computing $n+1$ scores (one for each ``key''), which then form probabilities. The outter dot-product multiplies each ``value'' by its probability and sums them to perform soft-selection.

\section{Experiments}\label{sec:experiments}

In this section, we study the following questions:
\begin{enumerate}[labelindent=0cm,labelsep=2pt,noitemsep,nolistsep, topsep=0pt,leftmargin=*]
  \setlength{\itemsep}{1pt}
  \setlength{\parskip}{0pt}
  \item Can \ourmodel~generalize memory to longer spatiotemporal combinations of object motions?
  \item  Can \ourmodel~generalize navigation towards and avoidance of 3D objects to larger environments?
  \item Can \ourmodel~generalize memory of goal-information to larger maps with more distractor-based obstacles?
\end{enumerate}

\begin{wraptable}[7]{r}{.4\textwidth}
  \vspace{-10pt}
  \begin{scriptsize}
  \caption{Baselines.}
  \vspace{-10pt}
  \begin{center}
    \begin{tabular}{l|cc}
      \toprule
        Method & \twoline{Observation}{Attention}  & \twoline{Modular}{State}  \\
      \midrule
      LSTM & \Xmark  & \Xmark \\ 
      AAA & Spatial  & \Xmark \\ 
      RIMs & Spatial & \Cmark \\ 
      \modelabb~(Ours) & Feature  & \Cmark \\
      \bottomrule
      \end{tabular}
  \end{center}
  \end{scriptsize}
  \label{table:classification}
\end{wraptable}
\textbf{Baselines.} Our first baseline is a common choice for learning state-representations, a \newterm{Long Short-term Memory (LSTM)}~\citep{hochreiter1997long}.
We study two other baselines that also attend to observation features: \newterm{Attention Augmented Agent (AAA)}~\citep{mott2019towards} and \newterm{Recurrent Independent Mechanisms (RIMs)}~\citep{goyal2019recurrent}.
Both employ transformer-style attention~\citep{locatello2020object, vaswani2017attention} to attend to individual \emph{spatial positions} by reducing observation features to a weighted average over spatial positions. 
We instead attend to \emph{features shared across all spatial positions}.
RIMs, like \modelabb, represents state with a set of recurrent modules. 
We expand on the differences between baselines in \S\ref{appendix:general-arch}.

\textbf{Implementation details.} 
We implement our recurrent observation encoder, $\phi$, as a ResNet~\citep{he2016deep} followed by a Convolutional LSTM (ConvLSTM)~\citep{shi2015convolutional}. 
We implement the update function of each module with an LSTM. 
We used multihead-attention~\citep{vaswani2017attention} for $\smash{f^{k}_{\tt share}}$. We trained the architecture with the IMPALA algorithm~\citep{espeholt2018impala} and an Adam optimizer~\citep{kingma2014adam}. 
We tune hyperparameters for all architectures with the ``Place X next to Y'' task from the BabyAI environment~\citep{babyai_iclr19} (\S~\ref{sec:experiments-distractors}). 
We expand on implementation details in \S\ref{appendix:implementation}.
For details on hyperparameters, see \S\ref{sec:hyperparameters}.

\subsection{Generalizing memory to more object motions}\label{sec:experiments-ballet}
We study this with the ``Ballet'' grid-world~\citep{lampinen2021towards} shown in ~\Figref{fig:environments} (a). 
\textbf{Tasks}. The agent controls a white square that begins in the middle of the grid. There are $m$ other ``ballet-dancer'' objects that move with a one of $15$ distinct object motions. The dances move in sequence for 16 time-steps with a 48-time-step delay in between. 
After all dancers finish, the agent is given a language instruction indicating the correct ballet dancer to navigate towards. 
All shapes and colors are randomized making motion the only feature indicating the goal object.
\textbf{Observations}. The agent observes a top-down RBG image of the environment.
\textbf{Actions}. The agent can move left, right, up, and down.
\textbf{Reward} is $1$ if it touches the correct dancer and $0$ otherwise.
\textbf{Tasks split}. Training tasks always consists of seeing $m=\{2,4\}$ dancers; testing tasks always consists of seeing $m=\{8\}$ dancers.
All agents learn with a sample budget of 2 billion frames.
A poorly performing agent will obtain chance performance, $1/m$.

\begin{figure*}[!htb]
  \input{text/plot-ballet.tex}
\end{figure*}
We present the training and generalization success rates in \Figref{fig:performance-ballet}.
We learned spatiotemporal observation features with RIMs and AAA for a fair comparison.
We found that only \modelabb~is able to obtain above chance performance for training and testing.
In order to understand the source of our performance, we ablate using a recurrent observation encoder, using multiple modules, and using feature-attention.
We confirm that a {recurrent encoder is required}.
Interestingly, we find that {either using multiple modules or using our feature-attention enables task-learning}, with our feature-attention mechanism being slightly more stable.

\subsection{Generalizing navigation with more 3D objects}\label{sec:experiments-playroom}
Here, we study the 3D Unity environment from~\citet{hill2020environmental} shown in~\Figref{fig:environments} (b). 
\textbf{Tasks}. The agent is an embodied avatar in a room filled with task objects and distractor objects. The agent receives a language instruction of the form ``\textit{X on Y}'' ---e.g., ``toothbrush on bed''.
We partition objects into two sets as follows: pickup-able objects $O_1=A \cup B$ and objects to place them on $O_2=C \cup D$.
\textbf{Observations.} The agent receives first-person egocentric RGB images. 
\textbf{Actions.} The agent has 46 actions that allow it to navigate, pickup and place objects. 
\textbf{Reward} is $1$ if it completes the task and $0$ otherwise. 
\textbf{Tasks split}. During training the agent sees $A \times D$ and $B \times C$ in a $4m \times 4m$ room with $4$ distractors, along with $A \times C$ and $B \times D$ in a $3m \times 3m$ room with $0$ distractors. We test the agent on $A \times C$ and $B \times D$ in a $4m \times 4m$ room with $4$ distractors.
We also train with ``Go to X'' and ``Lift X''.

\begin{figure*}[!htb]
  \input{text/plot-playroom.tex}
\end{figure*}
{We present the generalization success rate in \Figref{fig:performance-playroom}.} 
We find that baselines which used spatial attention learn more slowly than an LSTM or \modelabb. Additionally, both models that use spatial attention have poor performance until the end of training where AAA begins to improve. 
\modelabb~achieves relatively good performance, achieving a success rate of ~$60\%$ and ~$80\%$ on the two test settings, respectively.

\subsection{Generalizing to larger maps with more objects}\label{sec:experiments-keybox}

To study this, we create the ``keyBox'' environment depicted in \Figref{fig:environments} (c).
\textbf{Tasks} are defined with $n$ levels.
Each level is a hallway with a single box and a \emph{key of the same color} that the agent must retrieve.
The agent and the box always starts in the left-most end and the goal key always starts in the right-most end.
The agent always begins in the first level. It is teleported to the next level after placing the goal key next to the goal box.
The hallway for level $n$ consists of a length-$n$ sequence of $w \times w$ environment subsections.
Each subsection contains $d$ distractor objects.
\textbf{Observations}. The agent observes egocentric top-down images over a short segments of the hallway.
\textbf{Actions}. The agent can move forward, turn left, turn right, pick up objects and drop them.
\textbf{Rewards}. When completing a level, the agent gets a reward of $n/n_{\tt max}$ where $n_{\tt max}$ is the maximum level.
\textbf{Tasks split}.
Learning tasks include levels $1$ to $n_{\tt max}=10$. Test tasks only use levels $2n_{\tt max}$ and $3n_{\tt max}$.
{We study two generalization settings}: a \emph{densely populated setting} with subsections of area $w^2=9$ and $d=2$ distractors, and a \emph{sparsely populated setting} with subsections of area $w^2=25$ and $d=4$ distractors. 

\begin{figure*}[!htb]
  \centering
  \includegraphics[width=\textwidth]{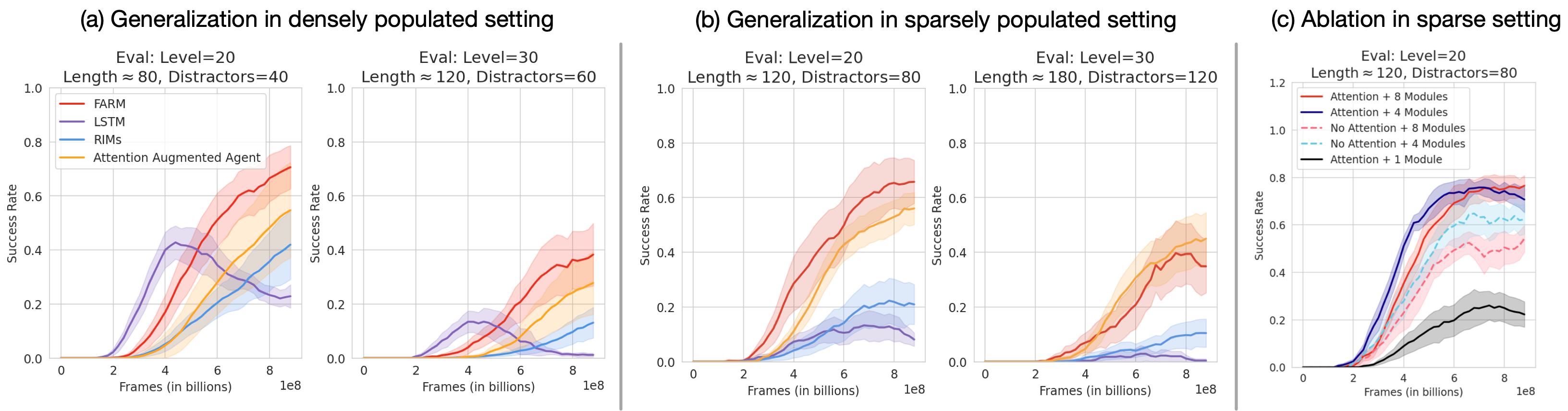}
  \caption{
    \textbf{\modelabb~enables generalizing memory of goal-information and avoidance of obstacles to larger maps with more objects}.
    We show the the success rate mean and standard error computed using 10 seeds.
    (a) In the densely populated setting, \modelabb~better generalizes to longer hallways with more distractors.
    (b) In the sparsely populated setting, \modelabb~has slightly better performance than AAA for $2n_{\tt max}$ but comparable performance for $3n_{\tt max}$.
    (c) Using multiple modules and feature attention both improve generalization.
    These results suggest that spatial attention interferes with generalization benefits of learning multiple modules. Learning feature attention and multiple modules, instead, act synergistically.
    }\label{fig:performance-keybox}
\end{figure*}

{We present the generalization success rates in \Figref{fig:performance-keybox}}.
In the dense setting, we see an LSTM quickly overfits in both settings.
All architectures with attention continue to improve in generalization performance as they continue training. 
In the dense setting, we find that \modelabb~ generalizes better (by about $20\%$ for AAA and about $30\%$ for RIMs).
In the sparse setting, both RIMs and an LSTM fail to generalize above $30\%$. \modelabb~generalizes better than AAA for level $20$ but gets comparable performance for level $30$.
In some ways, this is our most surprising result since it is not obvious that either learning multiple modules or using feature attention should help with this task.
In the next section we study possible sources of our generalization performance.

\subsubsection{Analysis of state representations} \label{sec:analysis}

\begin{figure*}[!htb]
  \centering
  \includegraphics[width=.9\textwidth]{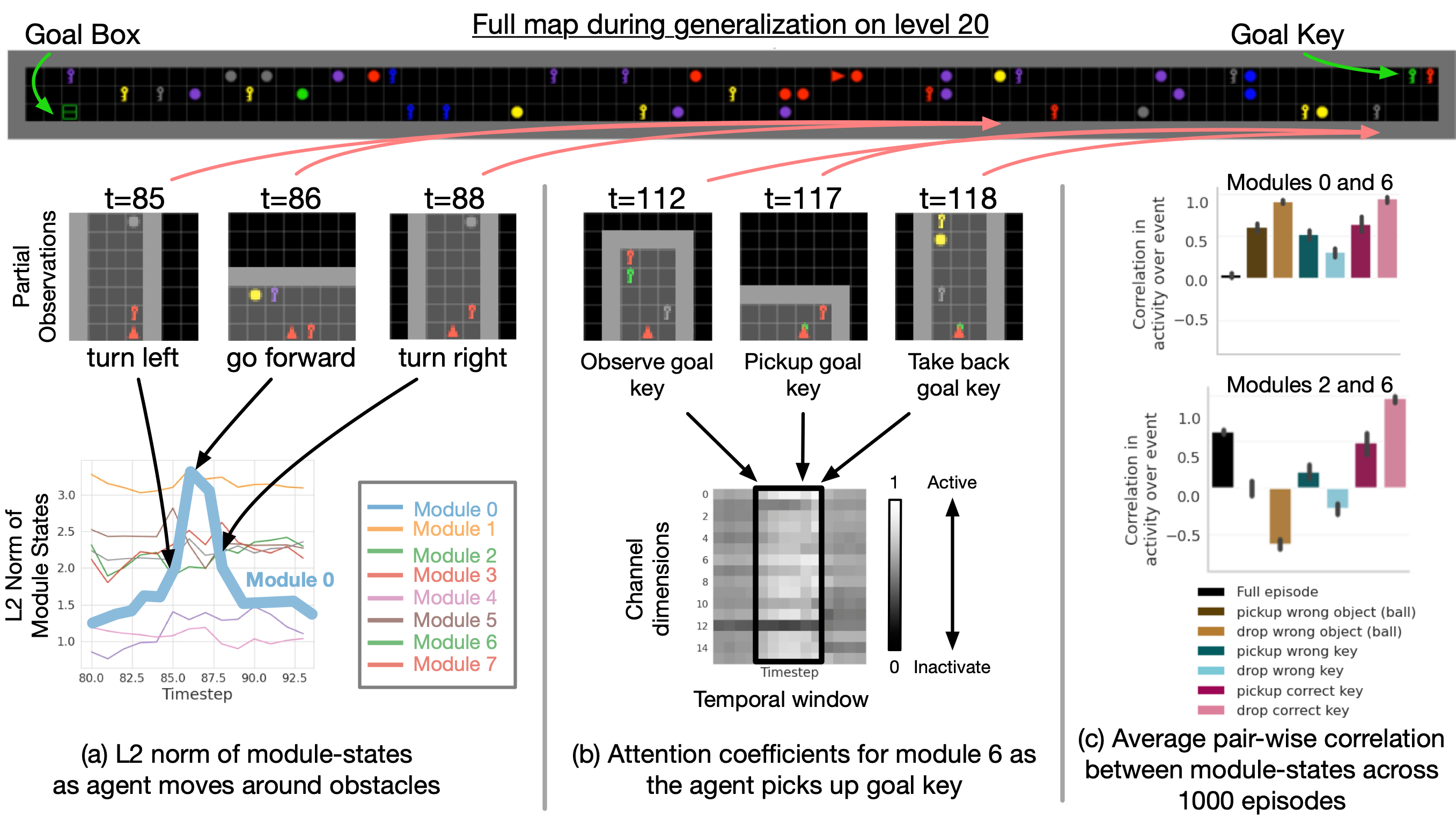}
  \caption{
    \textbf{We show evidence that different subsets of modules jointly represent object-induced task regularities.}
    (a) Module 0 commonly exhibits salient activity when the agent moves around an obstacle.
    (b) Module 6 activates its attention coefficients as the agent picks up the goal key.
    (c) Modules 2 and 6 correlate for picking up the correct key but anti-correlate for dropping the wrong object. 
    This is similar to when neurons in word embeddings correlate for some words (e.g. man and king), but anti-correlate for other words (e.g. man and woman).
    In general, we find rich patterns of correlation and anti-correlation between the modules.
    These results suggest that \modelabb~is representing task regularities across the modules in complicated and interesting ways. 
    Videos of the state-activity and attention coefficients: \href{https://bit.ly/3qCxatr}{https://bit.ly/3qCxatr}.
  }\label{fig:keybox-activity}
  \vspace{-10pt}
\end{figure*}

We study the state representations \modelabb~learns for categories of regularly occurring events.
We collect $2000$ generalization episodes in level $20$. We segment these episodes into $6$ categories: pickup ball, drop ball, pickup wrong key, drop wrong key, pickup correct key, and drop correct key.
We study the time-series of the L2 norm of each module-state and their attention coefficients.
For reference, we also show the L2 norm for the entire episode.
We note that we observed consistent activity that was not captured by our simple programmatic classification of states; for example, salient activity from module $0$ when the agent moved around obstacles. We show an example in \Figref{fig:keybox-activity} a.

Due to space constraints, we present a subset of results in \Figref{fig:keybox-activity}. For all results, please see the \S\ref{sec:full-analysis}.
While some modules are selective for different recurring events such as attending to goal information (\Figref{fig:keybox-activity}, b), it seems that subsets of modules jointly represent different aspects of state.
We hypothesize that this enables \modelabb~to leverage overlapping sets of modules to store goal-information or to navigate around obstacles in a decoupled way that supports recombination.
This is further supported by our ablation where we find that having $4$ or $8$ modules significantly outperforms using a single large module (all had about 8M params) (\Figref{fig:performance-keybox}, (c)). Feature attention consistently improves performance.

\section{Discussion and conclusion} \label{sec:discussion}

We have presented \modelabb, a novel state representation learning architecture for environments that have object-induced structural regularities.
Our results show that we can improves an RL agent's ability to generalize to out-of-distribution tasks by having multiple modules attend to spatiotemporal features with feature attention.
{Specifically this enables generalizing (a)  memory to longer combinations of object-motions (\S \ref{sec:experiments-ballet}); (b) navigation around 3D objects to larger environments (\S \ref{sec:experiments-playroom}); and (c) memory of goal information to longer sequences of obstacles (\S \ref{sec:experiments-keybox}). 
Our ablations suggest that feature attention mainly helps with long-horion memory. 
Interestingly, learning multiple modules helped across all conditions (memory, obstacle-avoidance, and language learning).
Our analysis suggests that learning multiple modules enables subsets to represent object-centric task-relevant events in flexible ways. We hypothesize that this  enables a deep RL agent to flexibly recombine its experience for generalization.

We compared FARM to other architectures that used spatial attention as a weak inductive bias for enabling objects to emerge in a state representation. We found that spatial attention hindered learning tasks with object motions and 3D objects. In the KeyBox task, spatial attention seemed to help AAA most in the sparse setting with many objects. This makes sense since spatial attention has been shown to help with distractors and the agent mainly needed to ignore objects and move forward. Interestingly, pairing spatial attention with multiple modules (RIMs) removed the benefits of both.
}

One limitation of FARM is that feature attention is not spatially invariant since it treat all positions as unique. Future work can look to adapt this attention for something that still describes multiple positions but in a spatially invariant way.
Another limitation of FARM is the length of temporal regularities it can capture.
Transformers~\citep{vaswani2017attention} have shown strong performance for representing long sequences. An interesting next-step might be to integrate \modelabb~with a transformer.
We hope that our work contributes to future RL algorithms that leverage weak inductive biases for capturing object-centric task regularities.

\bibliography{bib}

\begin{thebibliography}{42}
\providecommand{\natexlab}[1]{#1}
\providecommand{\url}[1]{\texttt{#1}}
\expandafter\ifx\csname urlstyle\endcsname\relax
  \providecommand{\doi}[1]{doi: #1}\else
  \providecommand{\doi}{doi: \begingroup \urlstyle{rm}\Url}\fi

\bibitem[Andreas et~al.(2016)Andreas, Rohrbach, Darrell, and
  Klein]{andreas2016neural}
Jacob Andreas, Marcus Rohrbach, Trevor Darrell, and Dan Klein.
\newblock Neural module networks.
\newblock In \emph{Proceedings of the IEEE conference on computer vision and
  pattern recognition}, pp.\  39--48, 2016.

\bibitem[Barreto et~al.(2020)Barreto, Hou, Borsa, Silver, and
  Precup]{barreto2020fast}
Andr{\'e} Barreto, Shaobo Hou, Diana Borsa, David Silver, and Doina Precup.
\newblock Fast reinforcement learning with generalized policy updates.
\newblock \emph{Proceedings of the National Academy of Sciences}, 117\penalty0
  (48):\penalty0 30079--30087, 2020.

\bibitem[Borsa et~al.(2018)Borsa, Barreto, Quan, Mankowitz, Munos, van Hasselt,
  Silver, and Schaul]{borsa2018universal}
Diana Borsa, Andr{\'e} Barreto, John Quan, Daniel Mankowitz, R{\'e}mi Munos,
  Hado van Hasselt, David Silver, and Tom Schaul.
\newblock Universal successor features approximators.
\newblock \emph{arXiv preprint arXiv:1812.07626}, 2018.

\bibitem[Bradbury et~al.(2018)Bradbury, Frostig, Hawkins, Johnson, Leary,
  Maclaurin, Necula, Paszke, Vander{P}las, Wanderman-{M}ilne, and
  Zhang]{jax2018github}
James Bradbury, Roy Frostig, Peter Hawkins, Matthew~James Johnson, Chris Leary,
  Dougal Maclaurin, George Necula, Adam Paszke, Jake Vander{P}las, Skye
  Wanderman-{M}ilne, and Qiao Zhang.
\newblock {JAX}: composable transformations of {P}ython+{N}um{P}y programs.
\newblock 2018.
\newblock URL \url{http://github.com/google/jax}.

\bibitem[Brooks et~al.(2021)Brooks, Rajendran, Lewis, and
  Singh]{brooks2021reinforcement}
Ethan~A Brooks, Janarthanan Rajendran, Richard~L Lewis, and Satinder Singh.
\newblock Reinforcement learning of implicit and explicit control flow in
  instructions.
\newblock \emph{ICML}, 2021.

\bibitem[Burgess et~al.(2019)Burgess, Matthey, Watters, Kabra, Higgins,
  Botvinick, and Lerchner]{burgess2019monet}
Christopher~P Burgess, Loic Matthey, Nicholas Watters, Rishabh Kabra, Irina
  Higgins, Matt Botvinick, and Alexander Lerchner.
\newblock Monet: Unsupervised scene decomposition and representation.
\newblock \emph{arXiv preprint arXiv:1901.11390}, 2019.

\bibitem[Carvalho et~al.(2021)Carvalho, Liang, Lee, Sohn, Lee, Lewis, and
  Singh]{carvalho2020reinforcement}
Wilka Carvalho, Anthony Liang, Kimin Lee, Sungryull Sohn, Honglak Lee,
  Richard~L Lewis, and Satinder Singh.
\newblock Reinforcement learning for sparse-reward object-interaction tasks in
  first-person simulated 3d environments.
\newblock \emph{IJCAI}, 2021.

\bibitem[Chaplot et~al.(2018)Chaplot, Sathyendra, Pasumarthi, Rajagopal, and
  Salakhutdinov]{chaplot2018gated}
Devendra~Singh Chaplot, Kanthashree~Mysore Sathyendra, Rama~Kumar Pasumarthi,
  Dheeraj Rajagopal, and Ruslan Salakhutdinov.
\newblock Gated-attention architectures for task-oriented language grounding.
\newblock In \emph{Proceedings of the AAAI Conference on Artificial
  Intelligence}, 2018.

\bibitem[Chevalier-Boisvert et~al.(2019)Chevalier-Boisvert, Bahdanau, Lahlou,
  Willems, Saharia, Nguyen, and Bengio]{babyai_iclr19}
Maxime Chevalier-Boisvert, Dzmitry Bahdanau, Salem Lahlou, Lucas Willems,
  Chitwan Saharia, Thien~Huu Nguyen, and Yoshua Bengio.
\newblock Baby{AI}: First steps towards grounded language learning with a human
  in the loop.
\newblock In \emph{International Conference on Learning Representations}, 2019.
\newblock URL \url{https://openreview.net/forum?id=rJeXCo0cYX}.

\bibitem[Diuk et~al.(2008)Diuk, Cohen, and Littman]{diuk2008object}
Carlos Diuk, Andre Cohen, and Michael~L Littman.
\newblock An object-oriented representation for efficient reinforcement
  learning.
\newblock In \emph{Proceedings of the 25th ICML}, pp.\  240--247, 2008.

\bibitem[D{\v{z}}eroski et~al.(2001)D{\v{z}}eroski, De~Raedt, and
  Driessens]{dvzeroski2001relational}
Sa{\v{s}}o D{\v{z}}eroski, Luc De~Raedt, and Kurt Driessens.
\newblock Relational reinforcement learning.
\newblock \emph{Machine learning}, 43\penalty0 (1):\penalty0 7--52, 2001.

\bibitem[Espeholt et~al.(2018)Espeholt, Soyer, Munos, Simonyan, Mnih, Ward,
  Doron, Firoiu, Harley, Dunning, et~al.]{espeholt2018impala}
Lasse Espeholt, Hubert Soyer, Remi Munos, Karen Simonyan, Vlad Mnih, Tom Ward,
  Yotam Doron, Vlad Firoiu, Tim Harley, Iain Dunning, et~al.
\newblock Impala: Scalable distributed deep-rl with importance weighted
  actor-learner architectures.
\newblock In \emph{ICML}, pp.\  1407--1416. PMLR, 2018.

\bibitem[Goyal et~al.(2020{\natexlab{a}})Goyal, Lamb, Gampa, Beaudoin, Levine,
  Blundell, Bengio, and Mozer]{goyal2020object}
Anirudh Goyal, Alex Lamb, Phanideep Gampa, Philippe Beaudoin, Sergey Levine,
  Charles Blundell, Yoshua Bengio, and Michael Mozer.
\newblock Object files and schemata: Factorizing declarative and procedural
  knowledge in dynamical systems.
\newblock \emph{arXiv}, 2020{\natexlab{a}}.

\bibitem[Goyal et~al.(2020{\natexlab{b}})Goyal, Lamb, Hoffmann, Sodhani,
  Levine, Bengio, and Sch{\"o}lkopf]{goyal2019recurrent}
Anirudh Goyal, Alex Lamb, Jordan Hoffmann, Shagun Sodhani, Sergey Levine,
  Yoshua Bengio, and Bernhard Sch{\"o}lkopf.
\newblock Recurrent independent mechanisms.
\newblock \emph{ICLR}, 2020{\natexlab{b}}.

\bibitem[Greff et~al.(2020)Greff, van Steenkiste, and
  Schmidhuber]{greff2020binding}
Klaus Greff, Sjoerd van Steenkiste, and J{\"u}rgen Schmidhuber.
\newblock On the binding problem in artificial neural networks.
\newblock \emph{arXiv}, 2020.

\bibitem[He et~al.(2016)He, Zhang, Ren, and Sun]{he2016deep}
Kaiming He, Xiangyu Zhang, Shaoqing Ren, and Jian Sun.
\newblock Deep residual learning for image recognition.
\newblock In \emph{CVPR}, pp.\  770--778, 2016.

\bibitem[Higgins et~al.(2017)Higgins, Pal, Rusu, Matthey, Burgess, Pritzel,
  Botvinick, Blundell, and Lerchner]{higgins2017darla}
Irina Higgins, Arka Pal, Andrei Rusu, Loic Matthey, Christopher Burgess,
  Alexander Pritzel, Matthew Botvinick, Charles Blundell, and Alexander
  Lerchner.
\newblock Darla: Improving zero-shot transfer in reinforcement learning.
\newblock In \emph{ICML}, pp.\  1480--1490. PMLR, 2017.

\bibitem[Hill et~al.(2020)Hill, Lampinen, Schneider, Clark, Botvinick,
  McClelland, and Santoro]{hill2020environmental}
Felix Hill, Andrew Lampinen, Rosalia Schneider, Stephen Clark, Matthew
  Botvinick, James~L McClelland, and Adam Santoro.
\newblock Environmental drivers of systematicity and generalization in a
  situated agent.
\newblock \emph{ICLR}, 2020.

\bibitem[Hochreiter \& Schmidhuber(1997)Hochreiter and
  Schmidhuber]{hochreiter1997long}
Sepp Hochreiter and J{\"u}rgen Schmidhuber.
\newblock Long short-term memory.
\newblock \emph{Neural computation}, 9\penalty0 (8):\penalty0 1735--1780, 1997.

\bibitem[Hu et~al.(2018)Hu, Shen, and Sun]{hu2018squeeze}
Jie Hu, Li~Shen, and Gang Sun.
\newblock Squeeze-and-excitation networks.
\newblock In \emph{Proceedings of the IEEE conference on computer vision and
  pattern recognition}, pp.\  7132--7141, 2018.

\bibitem[Jaderberg et~al.(2016)Jaderberg, Mnih, Czarnecki, Schaul, Leibo,
  Silver, and Kavukcuoglu]{jaderberg2016reinforcement}
Max Jaderberg, Volodymyr Mnih, Wojciech~Marian Czarnecki, Tom Schaul, Joel~Z
  Leibo, David Silver, and Koray Kavukcuoglu.
\newblock Reinforcement learning with unsupervised auxiliary tasks.
\newblock \emph{arXiv preprint arXiv:1611.05397}, 2016.

\bibitem[Justesen et~al.(2019)Justesen, Torrado, Bontrager, Khalifa, Togelius,
  and Risi]{justesen2018illuminating}
Niels Justesen, Ruben~Rodriguez Torrado, Philip Bontrager, Ahmed Khalifa,
  Julian Togelius, and Sebastian Risi.
\newblock Illuminating generalization in deep reinforcement learning through
  procedural level generation.
\newblock \emph{AAAI}, 2019.

\bibitem[Kabra et~al.(2021)Kabra, Zoran, Erdogan, Matthey, Creswell, Botvinick,
  Lerchner, and Burgess]{kabra2021simone}
Rishabh Kabra, Daniel Zoran, Goker Erdogan, Loic Matthey, Antonia Creswell,
  Matthew Botvinick, Alexander Lerchner, and Christopher~P Burgess.
\newblock Simone: View-invariant, temporally-abstracted object representations
  via unsupervised video decomposition.
\newblock \emph{arXiv preprint arXiv:2106.03849}, 2021.

\bibitem[Kansky et~al.(2017)Kansky, Silver, M{\'e}ly, Eldawy,
  L{\'a}zaro-Gredilla, Lou, Dorfman, Sidor, Phoenix, and
  George]{kansky2017schema}
Ken Kansky, Tom Silver, David~A M{\'e}ly, Mohamed Eldawy, Miguel
  L{\'a}zaro-Gredilla, Xinghua Lou, Nimrod Dorfman, Szymon Sidor, Scott
  Phoenix, and Dileep George.
\newblock Schema networks: Zero-shot transfer with a generative causal model of
  intuitive physics.
\newblock In \emph{ICML}, pp.\  1809--1818. PMLR, 2017.

\bibitem[Kingma \& Ba(2015)Kingma and Ba]{kingma2014adam}
Diederik~P Kingma and Jimmy Ba.
\newblock Adam: A method for stochastic optimization.
\newblock \emph{ICLR}, 2015.

\bibitem[Lampinen et~al.(2021)Lampinen, Chan, Banino, and
  Hill]{lampinen2021towards}
Andrew~Kyle Lampinen, Stephanie~CY Chan, Andrea Banino, and Felix Hill.
\newblock Towards mental time travel: a hierarchical memory for reinforcement
  learning agents.
\newblock \emph{arXiv}, 2021.

\bibitem[Lee et~al.(2020)Lee, Lee, Shin, and Lee]{Lee2020NetworkRA}
Kimin Lee, Kibok Lee, Jinwoo Shin, and Honglak Lee.
\newblock Network randomization: A simple technique for generalization in deep
  reinforcement learning.
\newblock In \emph{ICLR}, 2020.

\bibitem[Locatello et~al.(2020)Locatello, Weissenborn, Unterthiner, Mahendran,
  Heigold, Uszkoreit, Dosovitskiy, and Kipf]{locatello2020object}
Francesco Locatello, Dirk Weissenborn, Thomas Unterthiner, Aravindh Mahendran,
  Georg Heigold, Jakob Uszkoreit, Alexey Dosovitskiy, and Thomas Kipf.
\newblock Object-centric learning with slot attention.
\newblock \emph{arXiv preprint arXiv:2006.15055}, 2020.

\bibitem[Marom \& Rosman(2018)Marom and Rosman]{Marom2018ZeroShotTW}
Ofir Marom and Benjamin Rosman.
\newblock Zero-shot transfer with deictic object-oriented representation in
  reinforcement learning.
\newblock In \emph{NeurIPS}, 2018.

\bibitem[Mott et~al.(2019)Mott, Zoran, Chrzanowski, Wierstra, and
  Rezende]{mott2019towards}
Alex Mott, Daniel Zoran, Mike Chrzanowski, Daan Wierstra, and Danilo~J Rezende.
\newblock Towards interpretable reinforcement learning using attention
  augmented agents.
\newblock \emph{NeurIPS}, 2019.

\bibitem[Oh et~al.(2017)Oh, Singh, Lee, and Kohli]{Oh2017ZeroShotTG}
Junhyuk Oh, Satinder Singh, Honglak Lee, and P.~Kohli.
\newblock Zero-shot task generalization with multi-task deep reinforcement
  learning.
\newblock \emph{ICML}, abs/1706.05064, 2017.

\bibitem[Packer et~al.(2018)Packer, Gao, Kos, Kr{\"a}henb{\"u}hl, Koltun, and
  Song]{packer2018assessing}
Charles Packer, Katelyn Gao, Jernej Kos, Philipp Kr{\"a}henb{\"u}hl, Vladlen
  Koltun, and Dawn Song.
\newblock Assessing generalization in deep reinforcement learning.
\newblock \emph{arXiv}, 2018.

\bibitem[Perez et~al.(2018)Perez, Strub, De~Vries, Dumoulin, and
  Courville]{perez2018film}
Ethan Perez, Florian Strub, Harm De~Vries, Vincent Dumoulin, and Aaron
  Courville.
\newblock Film: Visual reasoning with a general conditioning layer.
\newblock In \emph{Proceedings of the AAAI Conference on Artificial
  Intelligence}, 2018.

\bibitem[Shi et~al.(2015)Shi, Chen, Wang, Yeung, Wong, and
  Woo]{shi2015convolutional}
Xingjian Shi, Zhourong Chen, Hao Wang, Dit-Yan Yeung, Wai-Kin Wong, and
  Wang-chun Woo.
\newblock Convolutional lstm network: A machine learning approach for
  precipitation nowcasting.
\newblock \emph{Advances in neural information processing systems}, 28, 2015.

\bibitem[Sohn et~al.(2018)Sohn, Oh, and Lee]{sohn2018hierarchical}
Sungryull Sohn, Junhyuk Oh, and Honglak Lee.
\newblock Hierarchical reinforcement learning for zero-shot generalization with
  subtask dependencies.
\newblock \emph{NeurIPS}, 2018.

\bibitem[Sohn et~al.(2021)Sohn, Woo, Choi, and Lee]{sohn2020meta}
Sungryull Sohn, Hyunjae Woo, Jongwook Choi, and Honglak Lee.
\newblock Meta reinforcement learning with autonomous inference of subtask
  dependencies.
\newblock \emph{ICLR}, 2021.

\bibitem[Tobin et~al.(2017)Tobin, Fong, Ray, Schneider, Zaremba, and
  Abbeel]{tobin2017domain}
Josh Tobin, Rachel Fong, Alex Ray, Jonas Schneider, Wojciech Zaremba, and
  Pieter Abbeel.
\newblock Domain randomization for transferring deep neural networks from
  simulation to the real world.
\newblock In \emph{IROS}, pp.\  23--30. IEEE, 2017.

\bibitem[Vaswani et~al.(2017)Vaswani, Shazeer, Parmar, Uszkoreit, Jones, Gomez,
  Kaiser, and Polosukhin]{vaswani2017attention}
Ashish Vaswani, Noam Shazeer, Niki Parmar, Jakob Uszkoreit, Llion Jones,
  Aidan~N Gomez, Lukasz Kaiser, and Illia Polosukhin.
\newblock Attention is all you need.
\newblock \emph{arXiv}, 2017.

\bibitem[Watters et~al.(2019)Watters, Matthey, Bosnjak, Burgess, and
  Lerchner]{watters2019cobra}
Nicholas Watters, Loic Matthey, Matko Bosnjak, Christopher~P Burgess, and
  Alexander Lerchner.
\newblock Cobra: Data-efficient model-based rl through unsupervised object
  discovery and curiosity-driven exploration.
\newblock \emph{arXiv preprint arXiv:1905.09275}, 2019.

\bibitem[Zhang et~al.(2018)Zhang, Sukhbaatar, Lerer, Szlam, and
  Fergus]{zhang2018composable}
Amy Zhang, Sainbayar Sukhbaatar, Adam Lerer, Arthur Szlam, and Rob Fergus.
\newblock Composable planning with attributes.
\newblock In \emph{ICML}, pp.\  5842--5851. PMLR, 2018.

\bibitem[Zhang et~al.(2021)Zhang, McAllister, Calandra, Gal, and
  Levine]{zhang2021learning}
Amy Zhang, Rowan McAllister, Roberto Calandra, Yarin Gal, and Sergey Levine.
\newblock Learning invariant representations for reinforcement learning without
  reconstruction.
\newblock \emph{ICLR}, 2021.

\bibitem[Zoran et~al.(2021)Zoran, Kabra, Lerchner, and Rezende]{zoran2021parts}
Daniel Zoran, Rishabh Kabra, Alexander Lerchner, and Danilo~J Rezende.
\newblock Parts: Unsupervised segmentation with slots, attention and
  independence maximization.
\newblock In \emph{Proceedings of the IEEE/CVF International Conference on
  Computer Vision}, pp.\  10439--10447, 2021.

\end{thebibliography}
\bibliographystyle{tmlr}

\clearpage
\appendix

\section{Full Correlations for Analysis}\label{sec:full-analysis}

\begin{figure}[!htb]
  \centering
  \begin{minipage}{.48\textwidth}
      \centering
      \includegraphics[width=\textwidth]{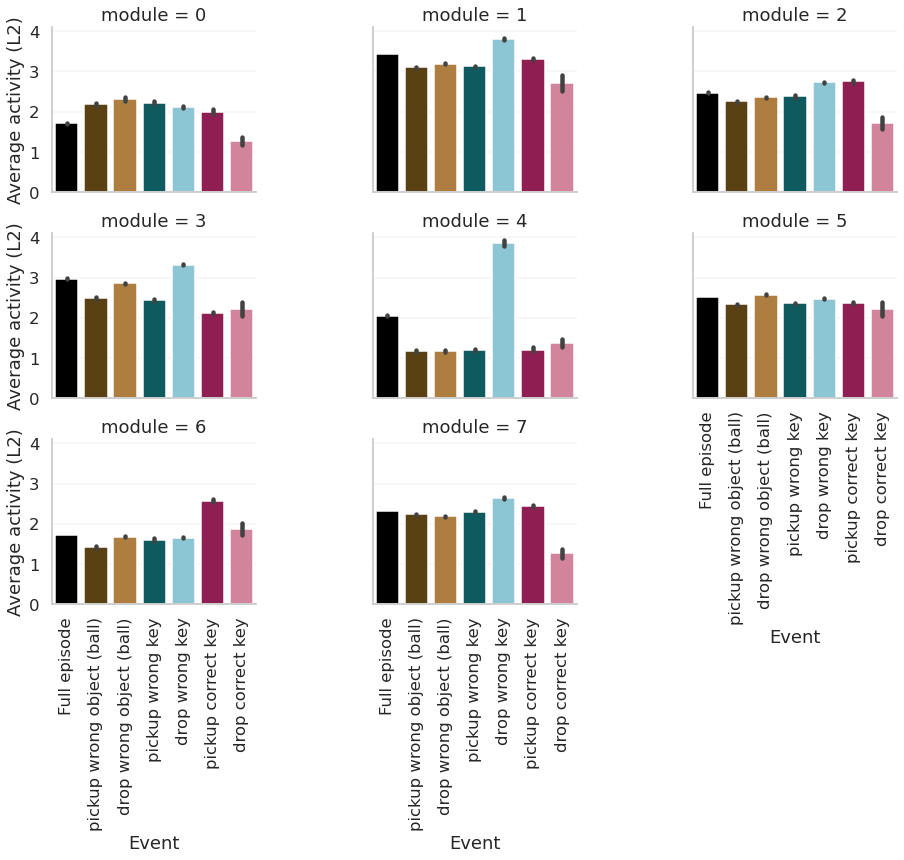}
      \caption{Trained weights.}\label{fig:activity-full-trained}
  \end{minipage}
  \hfill
  \begin{minipage}{.48\textwidth}
    \centering
    \includegraphics[width=\textwidth]{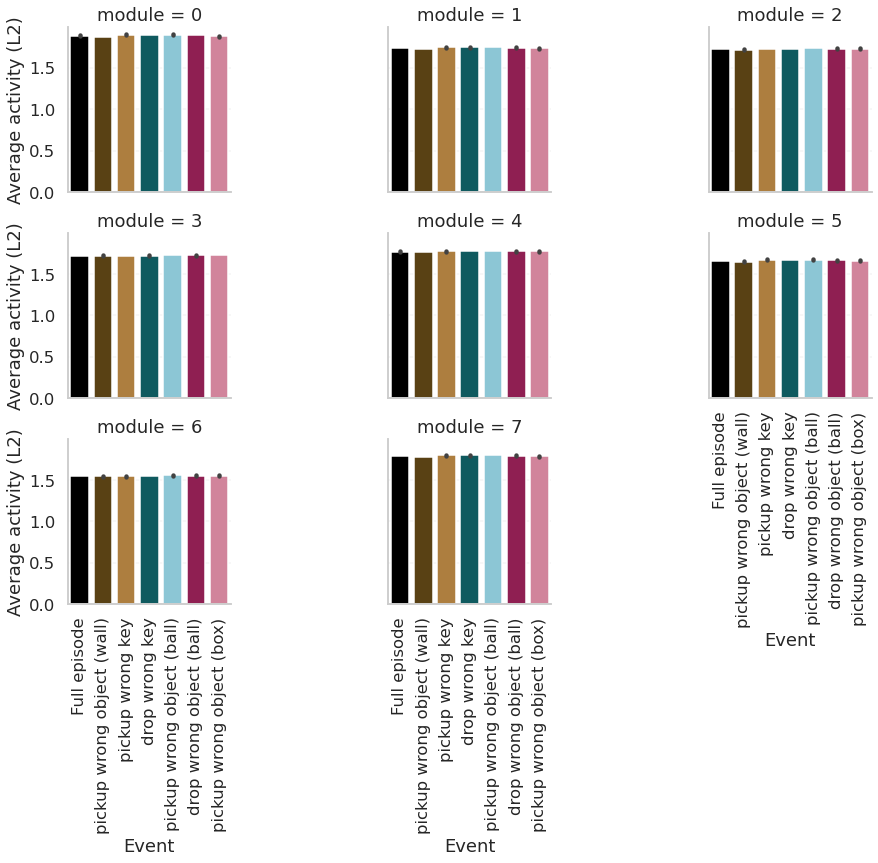}
    \caption{Random weights.}\label{fig:activity-full-random}  
  \end{minipage}
  \caption{Average L2 norm of module-states. We find that when weights are trained on the task, some modules are selective for different events. For example, Module 4 is selective for ``drop wrong key'' and module 6 is selective for ``pickup correct key''. When we use random weights, we see that all modules have the same activity for all events. This indicates that they have not learned any task-specific activity.}\label{fig:activity-full}
\end{figure}

\begin{figure}[!htb]
  \centering
  \includegraphics[width=\textwidth]{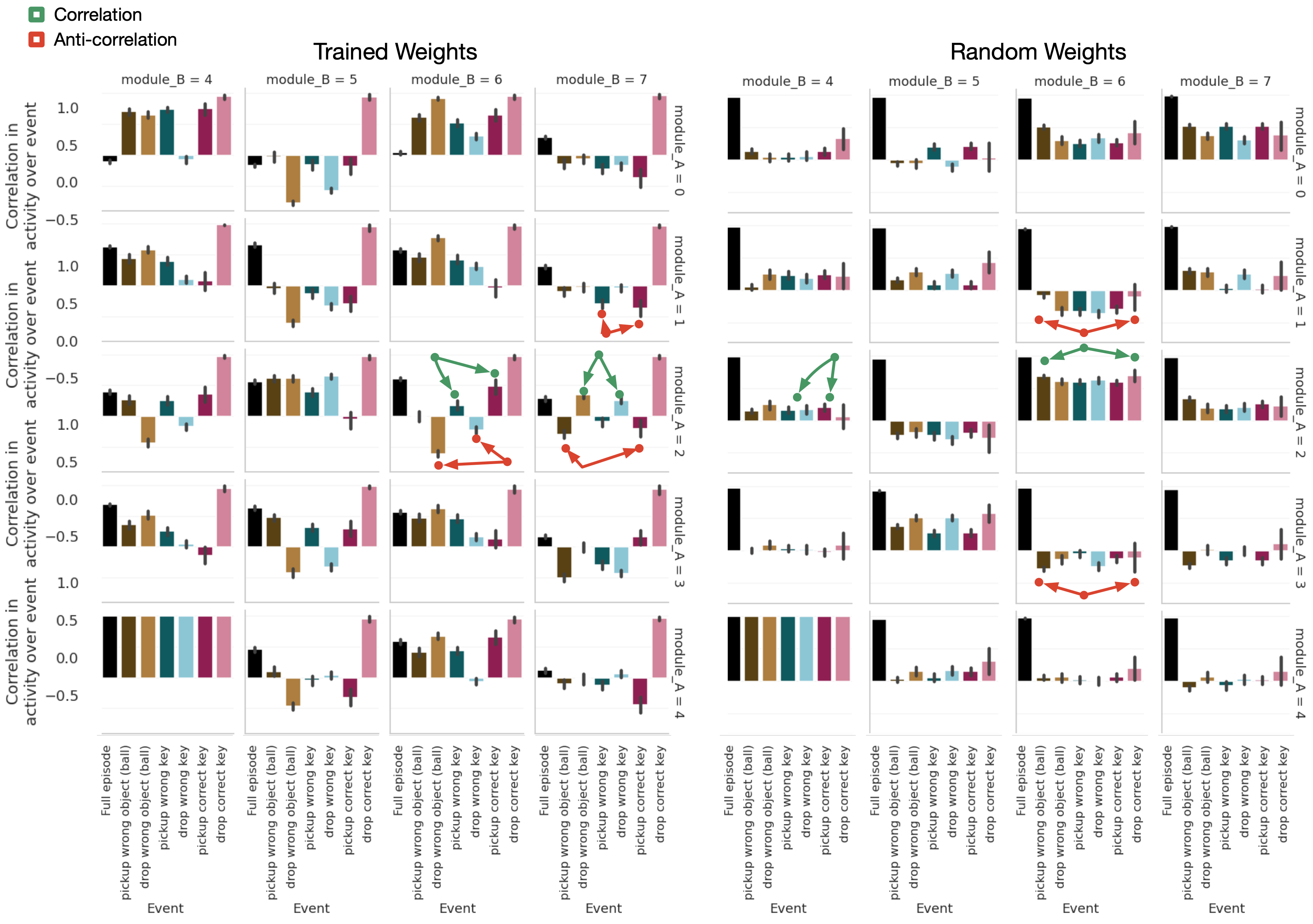}
  \caption{When looking at trained weights (left), we find that pairs of modules will have high correlation on some events and high anti-correlation on other events. For example,  modules 7 and 2 correlate for drop wrong object and drop wrong key but anti-correlate pickup wrong object and pickup correct key. If we look at random weights (right), we see that pairs of modules will either fully correlate (modules 6 and 2), fully anti-correlate (modules 6 and 1), or have weak/no correlation (modules 6 and 4) for events. Importantly, we don't see a significant mixture correlation and anti-correlation like we see with trained weights. This suggests that the random weights have less task-specific learning/uses by the agent.}\label{fig:correlation-comparison}
\end{figure}

\begin{figure}[!htb]
  \centering
  \includegraphics[width=\textwidth]{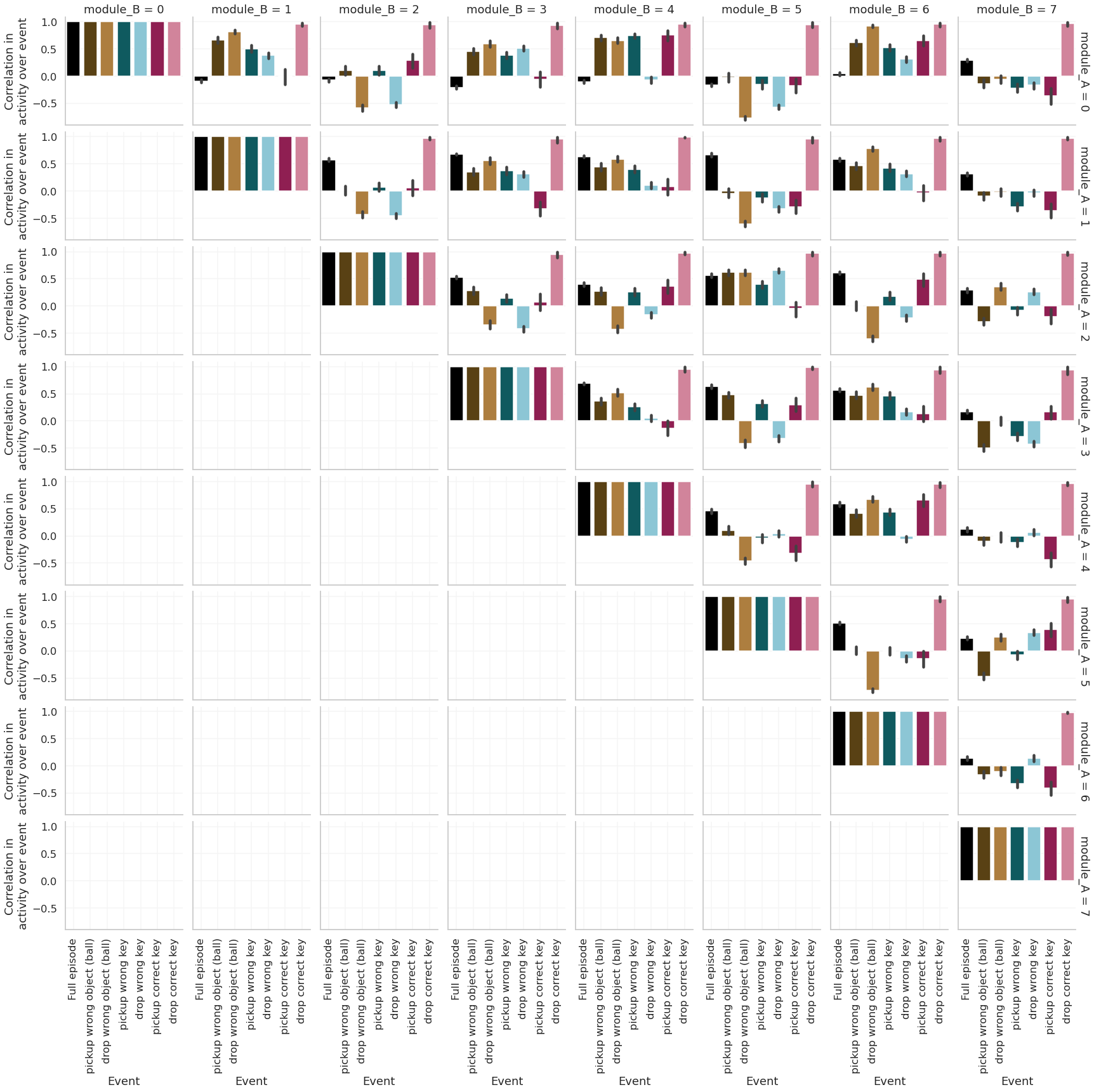}
  \caption{Average pair-wise correlation between L2 norm of module-states.}\label{fig:correlation-full}
\end{figure}

\begin{figure}[thb]
  \centering
  \includegraphics[width=.85\textwidth]{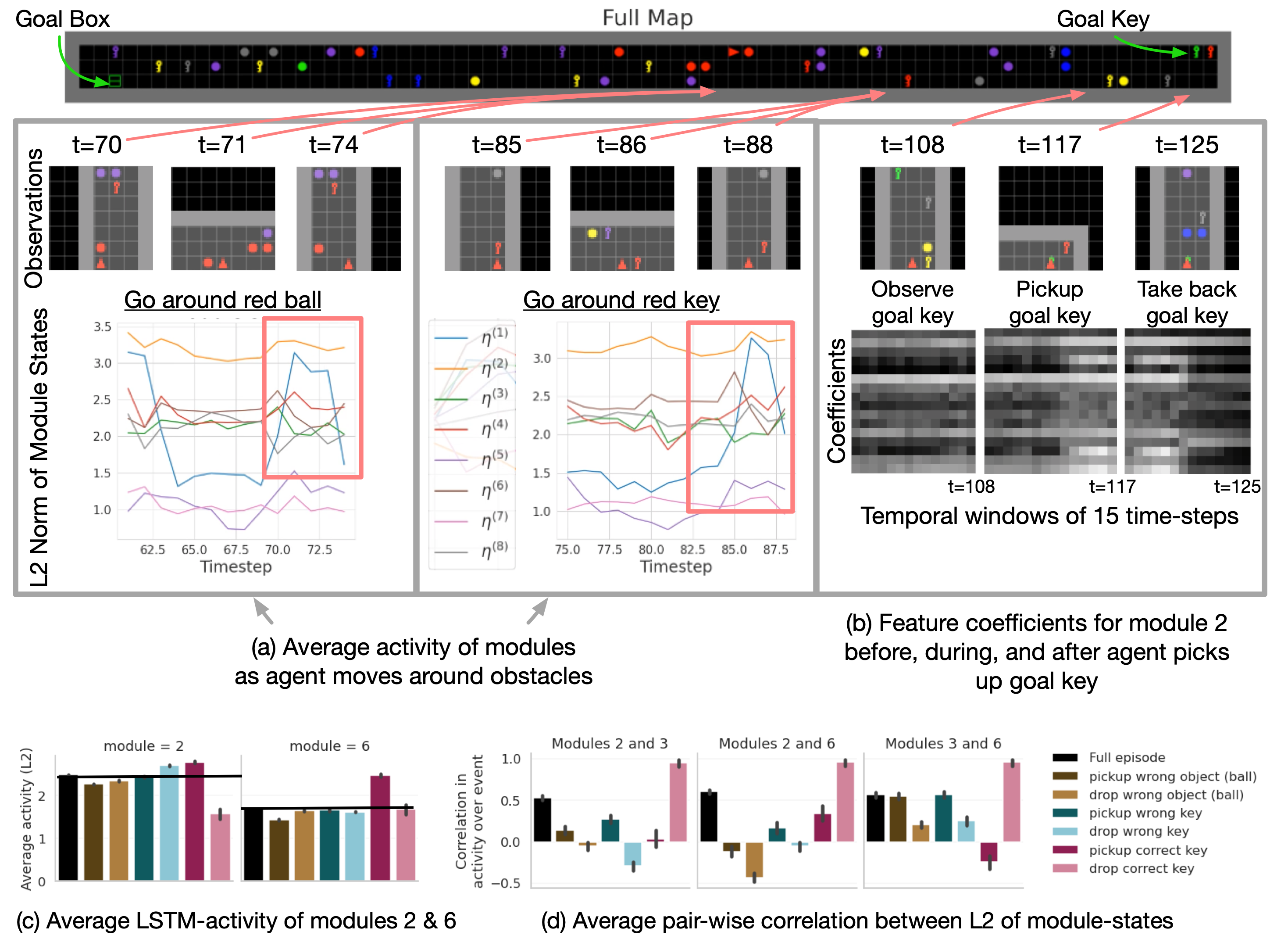}
  \caption{
    Panels (a) and (b) both show that different modules have selective activity on different events.
    (a) Module 0 exhibits salient activity when the agent moves around an obstacle.
    (b) Module 6 shows selective activity for representing goal information.
    (c) Module 6 also shifts its attention coefficients as the agent picks up the goal key.
    (d) We generally find that multiple modules activate for an event. Here, modules 3 and 6 show correlated activity for picking up a ball or non-goal key. 
    Videos of the state-activity and attention coefficients over test episodes: \href{https://bit.ly/3qCxatr}{https://bit.ly/3qCxatr}.
  }\label{fig:keybox-activity-full}
  \vspace{-10pt}
\end{figure}

In this section, we show full plots for the analysis in \S \ref{sec:analysis}. We show
\begin{enumerate}
  \item Average L2 norm of all module-states (trained and random weights) (\Figref{fig:activity-full}).
  \item Comparison of average pair-wise correlation of module-states for trained and random weights (\Figref{fig:correlation-comparison}).
  \item Average pair-wise correlation between L2 norm of all module-states (\Figref{fig:correlation-full}).
  \item More in-depth plots of activity and attention coefficients (\Figref{fig:keybox-activity-full}).
\end{enumerate}

\clearpage

\section{Additional Experiments}

\subsection{Generalizing memory-retention to novel spatial compositions of object-dynamics}\label{sec:experiments-ballet-parallel}
We use a variant of the the task in \S \ref{sec:experiments-ballet}. The main difference is that in this setting, the dancers dance in parallel as opposed to in sequence. This task is no longer a test of memory but only a test of whether the agent can recognize separate object-motions.

\begin{figure}[!htb]
  \centering
  \includegraphics[width=\textwidth]{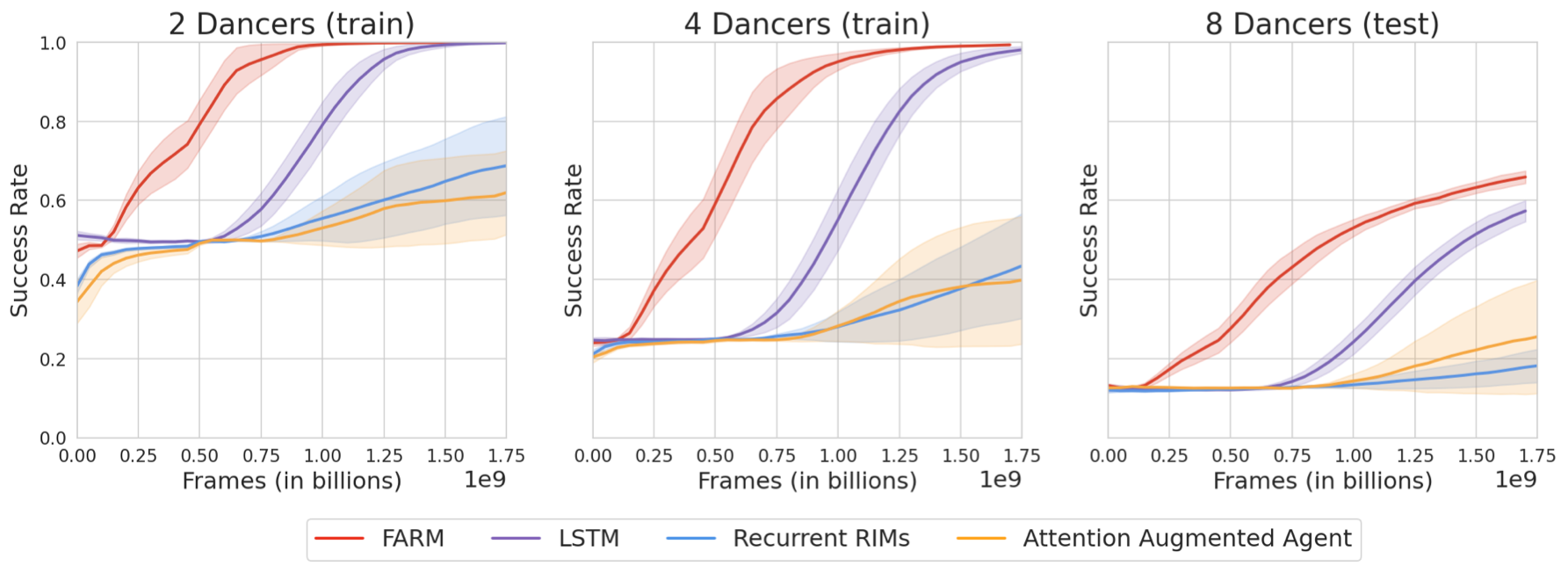}
  \caption{
    We present the success rate means and standard errors computed using 5 seeds.
    We find that \modelabb~more quickly learns and generalizes. The next best performance comes from using an LSTM. These results indicate that using spatial attention is an impediment to learning to recognize object-motions.
    }
  \label{fig:performance-ballet-parallel}
\end{figure}

{We present results in \Figref{fig:performance-putnext}}.
In the parallel dancing setting, we find that only \modelabb~and the LSTM can learn these tasks efficiently.
Both baselines that use spatial attention learn more slowly and with higher variance.

\subsection{Generalizing to an unseen number of distractors}\label{sec:experiments-distractors}

We study this with the ``Place $X$ next to $Y$'' task in the BabyAI gridworld~\citep{babyai_iclr19} (\Figref{fig:task-babyai-place}). 
The agent is a red triangle. Other objects can be squares, boxes or circles and they can take on 7 colors.
The agent receives a partial, egocentric observation of the environment (\Figref{fig:task-babyai-place}, right) and is given a synthetic language instruction.
The agent gets a reward of $1$ if chooses the correct dancer, and $0$ otherwise.
During training the agent sees either $0$ or $2$ distractors. During testing, the agent sees $11$ distractors.
As the number of distractors increases, the likelihood a distractor is either (a) confounding with the task objects or (b) blocks/confuses the agent also increases.

\begin{figure}[!htb]
  \centering
  \includegraphics[width=\textwidth]{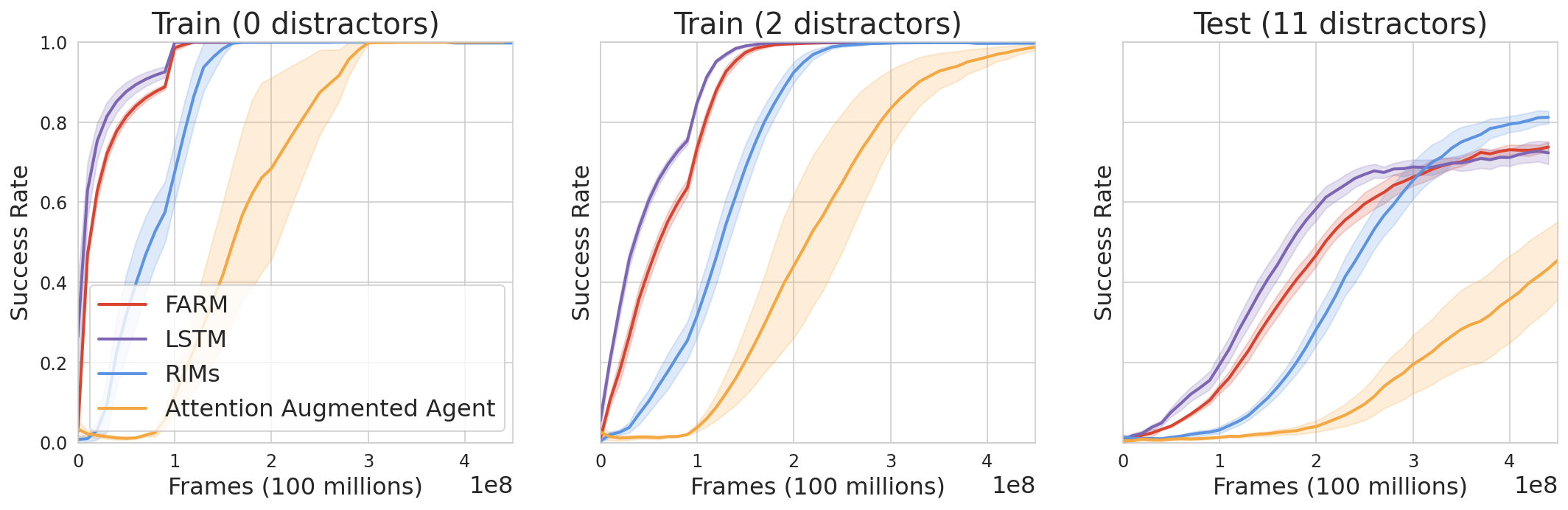}
  \caption{
    \textbf{RIMs, which uses spatial attention, better generalizes to more distractors.}
    We show train and test success rate performance for ``Place X next to Y'' in the BabyAI environment (10 runs).
    }
  \label{fig:performance-putnext}
\end{figure}

{We present results in \Figref{fig:performance-putnext}}.
On the left two panels, we present training results for $\{0,2\}$ distractors. All architectures can learn this task.
On the right-most panel, we present test results for $11$ distractors. \modelabb and an LSTM get comparable performance ($\approx 70\%$). RIMs has the best generalization success rate ($\approx 80\%$).

\clearpage

\section{Unified description of baseline methods}

\begin{figure}[bht]
\centering
\includegraphics[width=\textwidth]{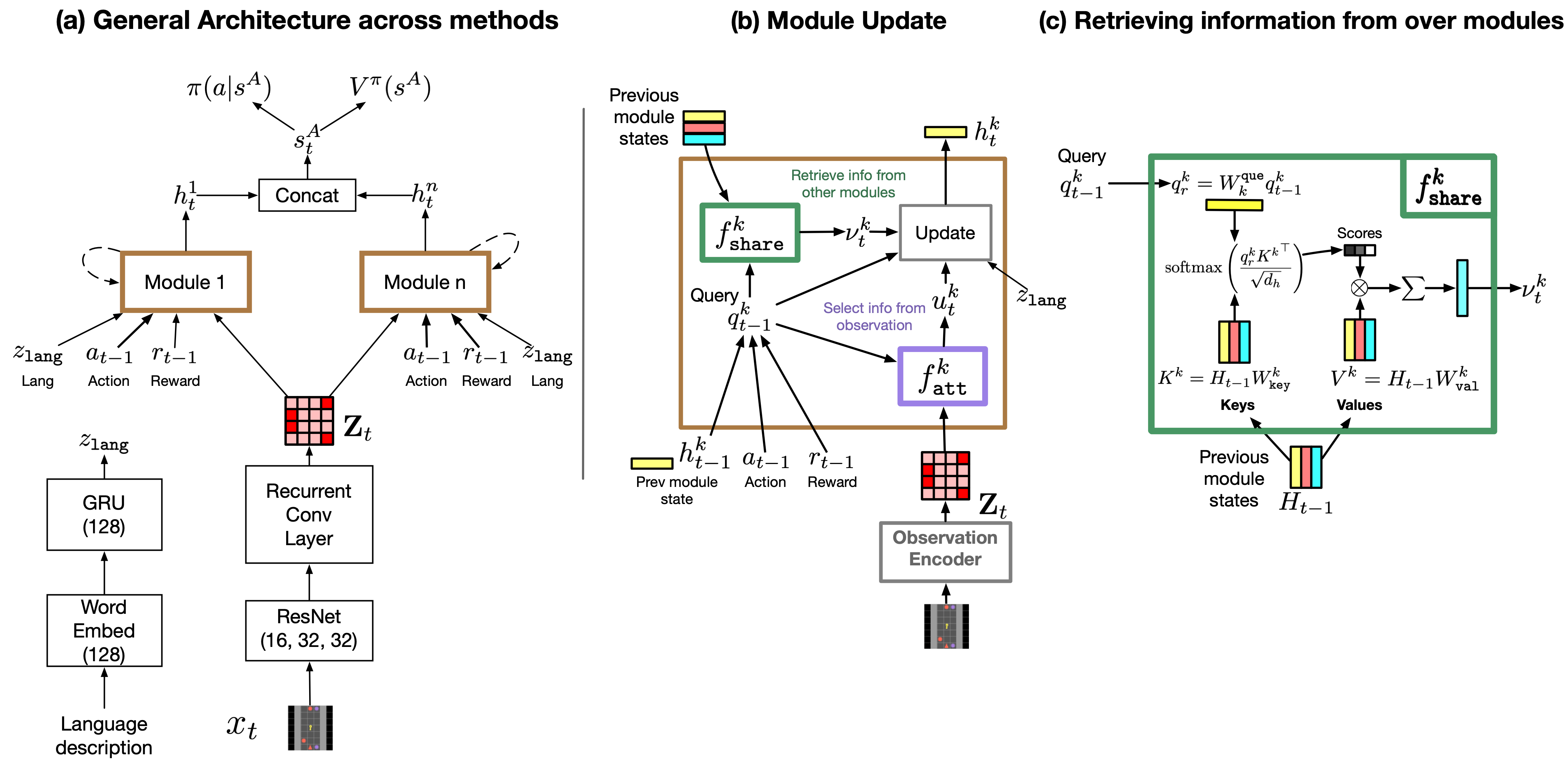}
\caption{
  \textbf{Schematic of architecture used by all attention-based methods}.
  The main difference between each method is in (a) the number of modules and (b) the attention function, $f^k_{\tt att}$, used to select information from the observation features.
  FARM uses $n$ modules and feature attention. RIMs uses $n$ modules and spatial attention. AAA uses $1$ module and spatial attention.
  An LSTM agent uses $1$ module that updates with (a) observation features (b) the language encoding, and (c) the previous reward and action. 
  }\label{fig:architecture-appendix}
\end{figure}

We present a detailed comparison of baseline methods.
In \Figref{fig:architecture-appendix}, we present a schematic of the general architecture that all methods used. The rest of this section is structured as follows. We first recap the general architecture used in all methods, which was describe in\S\ref{sec:architecture}. Both RIMs and FARM share their method for having modules share information. We describe this in \S\ref{appendix:general-arch-sharing}.
In \S\ref{appendix:general-arch-attn}, we describe spatial attention vs. feature attention.
In \S\ref{appendix:implementation}, we describe implementation details for these pieces.

\subsection{General Architecture}\label{appendix:general-arch}

At each time-step $t$, each module updates with both observation features and information from other modules.
First, the agent computes observation features with a recurrent observation encoder, $\mZ_{t} = \phi(o_t, \mZ_{t-1})$. 
Afterward, each module creates a \emph{query} vector by combining its previous module-state with the previous action and reward, $q^k_{t-1} = [h_{t-1}^{k}, a_{t-1}, r_{t-1}]$.
The query is used to attend to observation features via a dynamic feature attention mechanism $u^k_t = f^k_{\tt att}(\mZ_t, q^k_{t-1})$.
The query is also used to retrieve information from other modules with a transformer-style attention mechanism $\nu^k_t = f^k_{\tt share}(s^A_{t-1}, q^k_{t-1})$.
(We explain both attention mechanisms in more detail below).
Each module updates with both attention outputs to produce the next module-state $h_t^{k} = \eta^{k}(u^k_t, \nu^k_t, q^k_{t-1})$.
\revision{If a task additionally has a language description $o_{\tt lang}$ (as 2 of our experiments do), the module update also updates with an embedding of this description, $z_{\tt lang} = f_{\tt lang}(o_{\tt lang})$}.
Agent state is then defined by the combination of these module-states $s_t^A = [ h_t^{1}, \ldots, h_t^{\numschema}]$.
We summarize the computations below:
\begin{align}
  \mZ_t &= \phi(o_t, \mZ_{t-1}) && \text{obs features} \\
  q^k_{t-1} &=[h_{t-1}^{k}, a_{t-1}, r_{t-1}] && \text{query} \\
  u^k_t &= f^k_{\tt att}(\mZ_t, q^k_{t-1}) && \text{obs attention} \\
  \nu^k_t &= f^k_{\tt share}(s^A_{t-1}, q^k_{t-1}) && \text{share info} \\
  h_t^{k} &= \eta^{k}(u^k_t, \nu^k_t, q^k_{t-1}, z_{\tt lang}) && \text{module update} \\
  s_t^A &= [ h_t^{1}, \ldots, h_t^{\numschema}] && \text{agent state} 
\end{align}
\revision{where $[\cdot]$ is an operation that concatenates input vectors into a long vector.}

\subsubsection{Sharing information ($f^k_{\tt share}$)}\label{appendix:general-arch-sharing}
Both FARM and RIMs have modules that retrieves information from other modules using transformer-style attention~\citep{vaswani2017attention}. 
We define the collection of previous module-states as $\mH_{t-1} = \left[h_{t-1}^{(1)}; \ldots; h_{t-1}^{(n)}; \mathbf{0} \right] \in \mb{R}^{(n+1) \times d_h}$, where $\mathbf{0}$ is a null-vector used to retrieve no information. 
A module computes a ``retrieval query'' to search for information as $q_r^k=W^{\tt que}_{k} q^k_{t-1} \in R^{d_h}$.
That module computes ``retrieval keys and values'' as $K^k=\mH_{t-1} W^{\tt key}_{k} \in R^{n+1 \times d_h}$ and $V^k=\mH_{t-1} W^{\tt val}_{k} \in R^{n+1 \times d_h}$, respectively.
Each module then retrieves information as follows:
\begin{align}
  f^k_{\tt share}(s^A_{t-1}, q^k_{t-1}) = \softmax \left(
    \frac{
      q_r^k {K^k}^{\top}}{\sqrt{d_h}}\right) V^k.
\end{align}
Intuitively, the dot-product inside the softmax is computing $n+1$ scores (one for each ``key''), which then form probabilities. The outter dot-product multiplies each ``value'' by its probability and sums them to perform soft-selection.

\subsubsection{Observation attention ($f^k_{\tt attn}$)}\label{appendix:general-arch-attn}

\begin{figure}[thb]
  \centering
  \includegraphics[width=.9\textwidth]{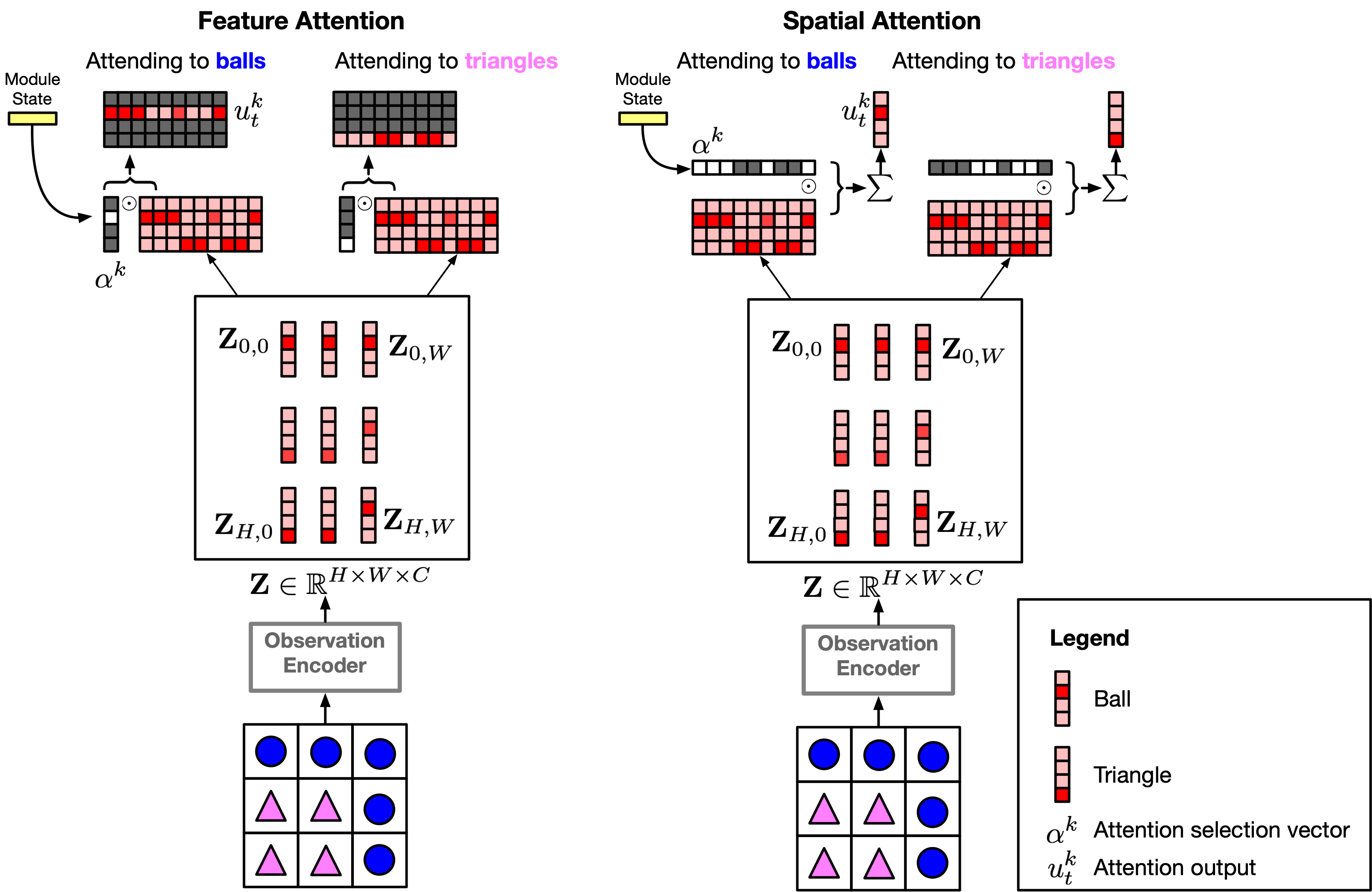}
  \caption{
    \textbf{Feature Attention vs. Spatial Attention}. We present a toy illustration of how modules may attend to different task-relevant objects. Our experiments indicate that modules respond to more abstract features (such as the presence of a general ``obstacle'' as in \Figref{fig:keybox-activity}). Still, it is illustrative to imagine how one could use spatial or feature attention to attend to either balls or triangles. When using feature attention, we get out a matrix that represents where balls or triangles are located across all positions. However, if we use spatial attention, features across spatial positions are averaged together. While the presence of the objects can be determined, their spatial location can be lost. While work can mitigate this by adding positional encodings, the resultant sum will have an average of relevant positional encodings. For the balls, the average of positional encoding may be a location that does not actually contain balls.
    }\label{fig:attention-appendix}
  \end{figure}

We present a diagram of feature attention vs. spatial attention in \Figref{fig:attention-appendix}. Below we describe updating with each type of attention. FARM uses feature attention. RIMs and AAA both use spatial attention. 

\textbf{Updating with Feature Attention}.
Here, state factors update with ``important`` features. We focused on visual features produced by a CNN, so this corresponds to important convolutional channels. This method essentially works by applying a learned mask to the convolutional features before updating with them.

Module $k$ transforms its query to a feature mask $\alpha^k$ by projecting the query and applying a sigmoid:
\begin{equation}
  \alpha^k  = \sigma(W^{\tt att}_{k} q^k_{t-1}) \mathbb{R}^{d_z}
\end{equation}
where $\sigma$ is a sigmoid function. Each dimension of the query is bounded between $0$ and $1$. This essentially gives an importance for updating with each of the $d_z$ features.
It then applies this mask to a projection of the convolutional features and then projects the masked features:
\begin{equation}
  u_t^k = W_2(\alpha^k \odot W_1 \mZ_t) \in \mathbb{R}^{m \times d_z}
\end{equation}

\textbf{Updating with Spatial Attention (used by RIMs and AAA)}.
Here, state factors update with spatial positions that contain relevant information.
A module computes a ``spatial query'' to search for observation information as 
\begin{equation}
  q_{\tt pos}^k=W^{\tt pos}_{k} q^k_{t-1}  \in R^{d_q}
\end{equation}
Observation features are then transformed to ``keys'' and ``values'' as 
$\mZ^{\tt key}=W^{\tt key}_k\mZ_t\in \mathbb{R}^{d_q \times m}$ and
$\mZ^{\tt val}=W^{\tt val}_k\mZ_t \in \mathbb{R}^{d_v \times m}$ (one for each spatial position). Each key is compared against the query, and the best match will be selected. 
First, ``soft'' selection scores $\alpha^k$ for each position are computed:
\begin{equation}
  \alpha^k = \softmax\left(\frac{q_{\tt pos}^k \mZ^{\tt key}}{\sqrt{d_q}}\right) \in \mathbb{R}^{m}
\end{equation}
One then obtains an update by doing a weighted sum over the values:
\begin{equation}
  u_t^k = \sum^m_{i=1} \alpha_i^k \mZ_{:,i}^{\tt val}
\end{equation}

\section{Implementation details}\label{appendix:implementation}
All neural networks were built using the Jax library~\citep{jax2018github}, haiku library~\citep{haiku2020github}, optax library~\citep{optax2020github}, and RLAX RL library. In all experiments, training was carried out using a distributed A3C setup~\citep{espeholt2018impala} with discrete actions. 
We trained all architectures end-to-end with the reinforcement learning objective via the IMPALA algorithm~\citep{espeholt2018impala} and an Adam optimizer~\citep{kingma2014adam}. 
For 3D Unity Env experiments, we added an additional Pixel Control loss~\citep{jaderberg2016reinforcement} for all agents. We used a single learner and 256 actors. 

\textbf{Observation encoder}.
We implement an agent's observation encoders, $\phi$, with a ResNet~\citep{he2016deep}. If the observation encoder is recurrent (as with FARM and AAA), the ResNet is followed by a Convolutional LSTM (ConvLSTM)~\citep{shi2015convolutional}. 
\textbf{Language encoder}.
Language descriptions are processed as follows. First, tokens are embedded into word embeddings and then they are fed into a GRU. The last token GRU embedding is used as the language description $z_{\tt lang}$.
\textbf{Module update}.
During update, modules (a) select information from other modules (RIMs, FARM) $\nu^k_t$ and (b) select observation information to update with $u^k_t$.
Modules then use an LSTM to update with the concatenation of $(\nu^k_t, u^k_t, a_{t-1}, r_{t-1}, h^k_{t-1})$, where $h^k_{t-1}$ is the modules' previous state.
\textbf{Sharing information}.
For both FARM and RIMs, we use used multihead-attention~\citep{vaswani2017attention} for sharing information, $\smash{f^{k}_{\tt share}}$ (see column c in \Figref{fig:architecture-appendix}).
For RIMs and AAA, we add positional emebddings for each spatial position of the convolutional features produced by the observation encoder.
\textbf{RL predictions}. Module states are then concatenated to form the agent's state representation, $s^A_t = [h^1_t, \ldots, h^n_t]$ and used to compute a policy $\pi(a|s^A_t)$ and estimate the state's value $V(s^A_t)$.

\clearpage

\section{Hyperparamters}\label{sec:hyperparameters}

Important training hyper-parameters are shown in Table~\ref{tab:rl-hyperparameters}, along with the components of the agent's architecture that are shared between the different models.
The parameter values used for each model presented in the main paper are shown below in Table~\ref{tab:model-params}.

Most hyperparameters (i.e.~for our RL algorithm, optimizer, and visual encoder) were tuned using a ``vanilla'' IMPALA agent that updated state using an LSTM. 
This is because all methods leveraged an LSTM to update state and we wanted to avoid bias towards our architecture. The only difference is that in AAA, there is one LSTM updating state and in RIMs and FARM, there are multiple LSTMs which are simultaneously being updated.

\subsection{Search on gridworld domains}
\textbf{Vanilla IMPALA LSTM agent}. We first searched RL algorithm (IMAPALA) and optimizer (Adam) hyperparameters with an LSTM on the ``Place X next to Y'' BabyAI task~\citep{babyai_iclr19}. We chose this task because our target domains were object-centric gridworlds and this simple object-centric grid-world acted as a sanity check that our methods worked.
We began with default values from our libraries and performed a random search using the following values: 
V-trace baseline cost $[1.0, .5, 0.1]$,
V-trace entropy cost $[10^{-2}, 10^{-3}, 10^{-4}]$,
V-trace $\gamma$ $[1.0, .99, .95]$, 
Adam learning rate $[10^{-3}, 5 \times 10^{-4}, 2 \times 10^{-3} 10^{-4}]$,
LSTM hidden size $[128, 256, 512]$. We consistently found that a larger memory had better results.

Once we found good IMPALA and Adam hyperparameters, we searched over agent-state hyperparameters for each method on the same BabyAI task. 
\textbf{\modelname}. We searched over attention projection dims $W_i \in [16, 32]$, Conv LSTM kernel size $[3, 5]$, and number of modules $[2, 4, 8]$. We set the ConvLSTM kernel size to be the same size as the final layer of the preceding ResNet. When using multihead attention, the number of attention relation heads is also a hyper-parameter. We fixed this to always be half off the number of modules. We used a per-module LSTM size of 128 and did not vary this across experiments.
\textbf{Attention Augmented agent}. We used hyper-parameters from their paper but tuned the following: LSTM hidden size $[256, 512]$, Attention query MLP size $[\{\null\}, \{256\}, \{256, 256\}]$, number of attention heads $[4, 8]$. We consulted the authors about our implementation.
\textbf{Recurrent Independent Mechanisms}. We used hyper-parameters from their paper but tuned the following: LSTM hidden size $[100, 128, 256]$, Observation/communication head size $[32, 64, 128]$, number of observation/communication heads $[4, 5, 6]$, number of RIMs $[4, 6, 9, 12]$. We consulted the authors about our implementation and used their source code for replication.

Finally, once we had good hyperparameters for agent-state, we applied the architectures to the ``Ballet'' and ``Keybox'' gridworld domains and explored whether increasing agent-state capacity (e.g. LSTM size or number of LSTMs) improved performance. We tried combinations of LSTM size and number of LSTMs that led each method to have approximately the same number of parameters. This was to ensure that no method performed better than the other simply because it had more parameters.

\subsection{Search on 3D unity domain} We recompleted our initial search on the RL algorithm (IMAPALA) and optimizer (Adam) hyperparameters. We searched over the same values as before and additionally searched over a larger MLPs for the policy and value heads $[200, 512]$, Adam optimizer episilon $[10^{-7}, 5 \times 10^{-8}, 10^{-8}]$,
Adam $\beta_1 [0.0, .9, .95, .99, .999]$ and $\beta_2 [0.0, .9, .95, .99, .999]$,
and did a small search over the Pixel Control loss scaling $[0.1, 0.01, 0.001]$ and Pixel Control discount factor $[0.9, .99]$. After we searched the IMPALA, Adam, and Pixel Control hyperparameters, we searched over individual architecture hyperparameters again.

\begin{table*}[t]
  \begin{center}
    \caption{\label{tab:rl-hyperparameters} Training hyper-parameters and shared network components used in experiments.}
    \begin{tabular}{lll}
    \toprule
    \textbf{Loss Hyper-parameters} & \textbf{3D Unity Env} & \textbf{Gridworlds}\\
    \midrule
    V-trace baseline cost & 1.0 & 0.5\\
    V-trace entropy cost & $10^{-4}$ & 0.01 \\
    V-trace $\gamma$ & 0.95 & 1.0 \\
    V-trace loss scaling & 0.1 & 1.0\\
    Pixel Control loss scaling & 0.1 & -- \\
    Pixel Control loss cell size & 4 & -- \\
    Pixel Control discount factor & 0.9 & -- \\
    Optimizer & clipped Adam & clipped Adam\\
    Learning rate & $2 \times 10^{-4}$ &  $10^{-4}$\\
    Max gradient Norm & 40.0 & 40.0\\
    Optimizer epsilon & $5 \times 10^{-8}$ & $10^{-8}$ \\
    Adam $\beta_1$ & 0.0 & 0.9\\
    Adam $\beta_2$ & 0.95 & 0.999\\
    \midrule
    \textbf{Shared Network Components} & & \\
    \midrule
    Language encoder & GRU & GUR \\
    Language encoder hidden sizes & 128 & 128 \\
    Language word embedding size & 128 & 128 \\
    Image encoder & Res-Net & Res-Net \\
    Res-Net channels & (16, 32, 32) & (16, 32, 32) \\
    Res-Net residual blocks & (2, 2, 2) & (2, 2, 2) \\
    Res-Net stride & 2 & 2 \\
    Res-Net kernel size & 3 & 3 \\
    Res-Net padding & SAME & SAME \\
    Image-language-reward-action combination & Concatenation & Concatenation \\
    Policy Head MLP shapes & [512, 46] & [200, 7] \\
    Value Head MLP shapes & [512, 1] & [200, 1] \\
    \bottomrule
    \end{tabular}
  \end{center}
\end{table*}

\begin{table*}[t]
  \begin{center}
    \caption{\label{tab:model-params} Model specific parameters. We highlight values that changed across environments in blue.}
    \begin{tabular}{llll}
    \toprule
    \textbf{Model parameter} & \textbf{3D Unity Env} & \textbf{\twoline{Gridworld}{Ballet}} & \textbf{\twoline{Gridworld}{Keybox}}\\
    \midrule
    \textbf{Observation Dims} & $72\times96$ & $99\times99$ & $56\times56$ \\
    \midrule
    \textbf{Feature-Attending Recurrent Modules} & & & \\
    \midrule
    Parameters (millions) & 5.1 &  7.1 &  7.6 \\
    {\color{blue}{Number of modules}} & 4 & 4 & 8 \\
    Module-state LSTM size & 128 & 128 & 128 \\
    {$\frac{\text{Relation heads}}{\text{Number of modules}}$} & .5 & .5 & .5 \\
    Projection dims $W_1, W_2$ & 16 & 16 & 16 \\
    ConvLSTM kernel size & 3 & 3 & 3 \\
    ConvLSTM hidden size & 32 & 32 & 32 \\
    \midrule
    \textbf{LSTM} & & & \\
    \midrule
    Parameters (millions) & 5.6 &  7.2 &  7.6 \\
    {\color{blue}{LSTM size}} & 896 & 768 & 1024 \\
    \midrule
    \textbf{Attention Augmented Agent} & & & \\
    \midrule
    Parameters (millions) & 5.1 &  6.9 &  7.5 \\
    ConvLSTM kernel size & 3 & 3 & 3 \\
    ConvLSTM output size & 128 & 128 & 128 \\
    {\color{blue}{LSTM size}} & 704 & 512 & 960\\
    Number of attention heads & 4 & 4 & 4\\
    Attention query MLP size & (256, 256) & (256, 256) & (256, 256) \\
    Positional basis dim & 4 & 4 & 4 \\
    \midrule
    \textbf{RIMs} & & & \\
    \midrule
    Parameters (millions) & 5 &  6.6 &  7.6 \\
    {\color{blue}{Number of modules}} & 12 & 9 & 9\\
    {{LSTM size}} & 128 & 128 & 128\\
    Observation heads & 6 & 6 & 6 \\
    Communication heads & 6 & 6 & 6\\
    Observation head size & 32 & 32 & 32 \\
    Communication head size & 32 & 32 & 32 \\
    Basis size & 4 & 4 & 4 \\
    Dropout & 0.2 & 0.2 & 0.2\\
    \bottomrule
    \end{tabular}
  \end{center}
\end{table*}
\clearpage

\section{Environments}

\begin{figure}[!htb]
  \centering

  \begin{minipage}{.65\textwidth}
      \centering
      \includegraphics[width=.78\textwidth]{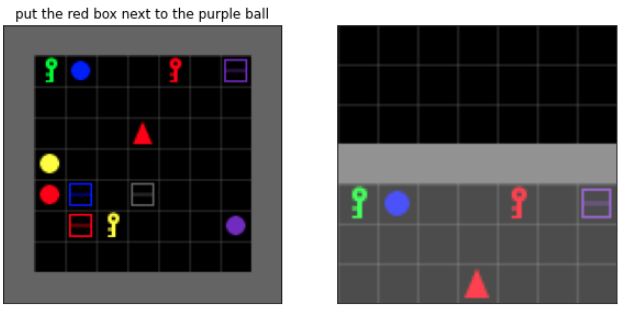}
      \caption{Place X on Y task in BabyAI environment.}
      \label{fig:task-babyai-place}
  \end{minipage}

  \hfill

  \begin{minipage}{.65\textwidth}
    \centering
    \includegraphics[width=.78\textwidth]{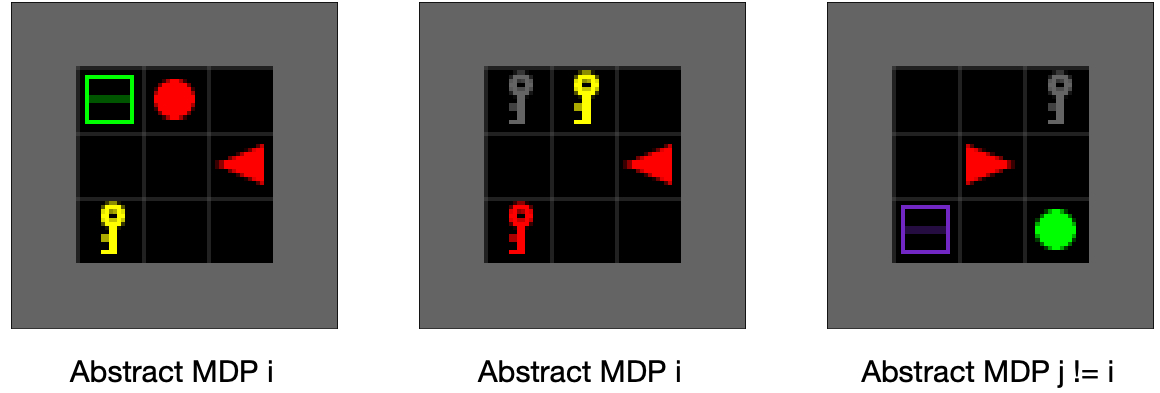}
    \caption{Abstract MDP Environment based on BabyAI.}
    \label{fig:task-babyai-abstract}    
  \end{minipage}

  \hfill

  \begin{minipage}{.65\textwidth}
    \centering
    \includegraphics[width=.78\textwidth]{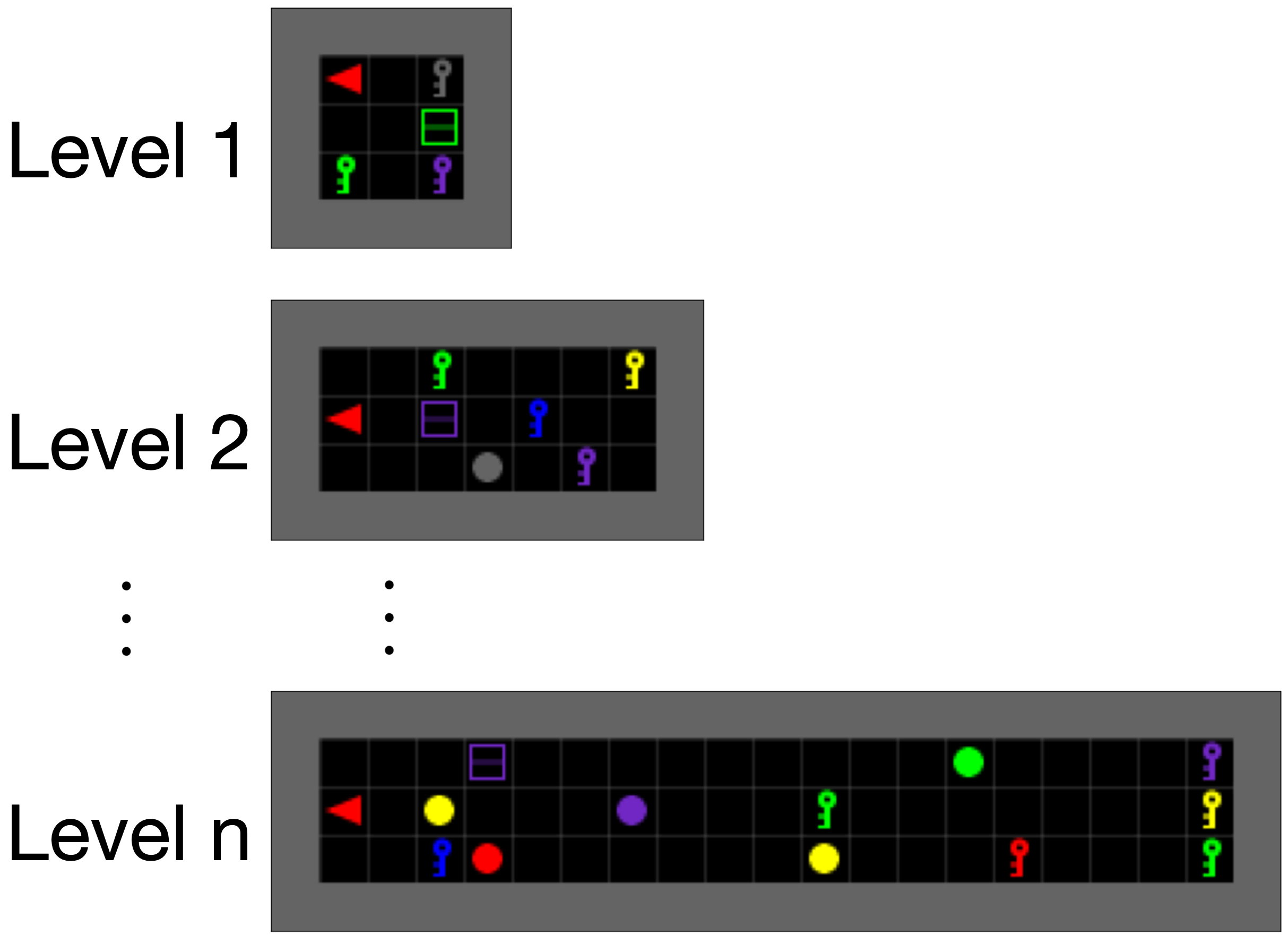}
    \caption{KeyBox task.}
    \label{fig:task-keybox}    
  \end{minipage}

  \hfill

  \begin{minipage}{.49\textwidth}
    \centering
    \includegraphics[width=.48\textwidth]{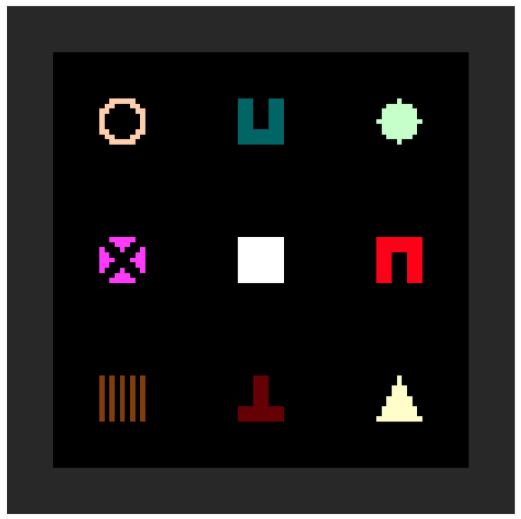}
    \caption{Ballet task.}
    \label{fig:task-ballet}    
  \end{minipage}

  \caption{Additional Environment Images.}
  \label{fig:appendix-envs}
\end{figure}

\subsection{Ballet}

Please refer to \citet{lampinen2021towards} for details on this task. Our only difference was to use tasks with $\{2,4\}$ dancers during training and tasks with $8$ dancers for testing.

\subsection{KeyBox}

\textbf{Observation Space}. The agent receives a $56 \times 56$ partially observable, egocentric image of the environment as in \Figref{fig:task-babyai-place}, right.

\textbf{Action Space}.
The action space is composed of the 7 discrete actions \textit{turn left}, \textit{turn right}, \textit{go forward}, \textit{pickup object}, \textit{drop object}, \textit{toggle}, and \textit{done/no-op}. 

\textbf{Reward function}. 
When the agent completes level $n$, it gets a reward of $n/n_{\tt max}$ where $n_{\tt max}$ is the maximum level the agent can complete.
We set $n_{\tt max}=10$ during training.
The agent has $50n$ time-steps to complete a level.

\begin{table}[htb]
  \begin{center}
    \caption{\label{tab:keybox-object-sets} Object and colors available for objects in the KeyBox task.}
    \begin{tabular}{ll}
    \toprule
    Set & Contains \\
    \midrule
    Shapes & ball, key, box \\
    \midrule
    Colors & red, green, blue, purple, pink, yellow, white \\
    \end{tabular}
  \end{center}
\end{table}

\subsection{3d Unity Environment} \label{sec:playroom-train-details}

For the ``place X on Y'' experiments in 3D, all pickupable objects were split into two sets $O_1 = A \cup B$ and all object to place something on into another two sets $O_2 = C \cup D$, as shown in Table \ref{tab:playroom-object-sets}. 
Given the challenging nature of the 3D environment (huge number of possible states, partial observability, language commands, long credit assignment), we had to employ a set of curriculum tasks in order for the agents to make any progress on the actual task of interest ``Put X on Y''.
The agent co-trained on the full set of tasks. This was possible since we used a distributed A3C setup for our training~\citep{espeholt2018impala}, where each of the actors generating the experience was running on one of the possible training levels. The different training tasks used during training and evaluation are shown in Table \ref{tab:playroom-curriculum}.

All episodes lasted for a maximum of 120 seconds and an action repeat of 4 was used. 
The images observations were rendered at $96 \times 72 \times 3$ and given to the agent along with a text language instruction, where each word in the instruction was mapped into a continuous vector of size $128$ using a fixed vocabulary of maximum size $1000$. 

\textbf{Reward function}. An agent get's a reward of $1$ if it completed the task and $0$ otherwise.

\textbf{Action Space}. The action space for the experiments in 3d Unity Environment was 46 discrete actions that allow the agent to move its body and change its head direction, to grab objects while moving and manipulate the held objects by rotating, pulling or pushing the held object.
The object is while as long as the agent is emitting a GRAB action, and dropped in the first instance that a GRAB action is not emitted. The full list of possible actions in the 3d Unity Environment environment is presented in Table \ref{tab:playroom-action-space}.

\begin{table}[htb]
  \begin{center}
    \caption{\label{tab:playroom-object-sets} Object and color set splits for the 3d Unity Environment ``Put X on Y'' experiments.}
    \begin{tabular}{ll}
    \toprule
    Set & Contains \\
    \midrule
    Set A (pickupable objects) & toilet roll, toothbrush, toothpaste \\
    Set B (pickupable objects) & bus, car, carriage, helicopter, keyboard \\
    Set C (support object) & stool, tv cabinet, wardrobe, wash basin \\
    Set D (support object) & bed, book case, chest, dining, table \\
    \midrule
    Colors & red, green, blue, aquamarine, magenta, orange, \\
           & purple, pink, yellow, white \\
    \end{tabular}
  \end{center}
\end{table}

\begin{table}[htb]
  \begin{center}
    \caption{\label{tab:playroom-curriculum} Descriptions of all the tasks used during training and evaluation. D refers to number of distractors and S to the room size.}
    \begin{tabular}{llll}
    \toprule
    Task name & S & D & Description \\
    \midrule
    \midrule
    Find X           & $4 \times 4$ & 5 & The agent is spawned randomly. \\
    (Set $A$ or $B$) & & & Room has $3$ objects from Set $A$ (or $B$) and $3$ from \\
                     & & & $C \cup D$ and instructed to go to an object from Set $A$ (or $B$). \\
                     & & & The purpose of these training tasks is to associate objects \\
                     & & & from Set $A$ and $B$ with their names and the ``find'' \\ 
                     & & & instruction with finding them. \\
    \midrule

    Find Y           & $4 \times 4$ & 5 & The agent is spawned randomly. \\
    (Set $C \cup D$) & & & Room has $3$ objects from Set $A$ (or $B$) and $3$ from \\
                     & & & $C \cup D$ and instructed to go to an object from Set $C \cup D$. \\
                     & & & The purpose of these training tasks is to associate objects \\
                     & & & from Set $C \cup D$ with their names and the ``find'' \\ 
                     & & & instruction with finding them. \\
    \midrule

    Lift X           & $4 \times 4$ & 5 & The agent is spawned randomly. \\
    (Set $A$ or $B$) & & & Room has $3$ objects from Set $A$ (or $B$) and $3$ from \\
                     & & & $C \cup D$ and instructed to lift an object from Set $A$ (or $B$). \\
                     & & & The purpose of these training tasks is to associate the ``lift'' \\ 
                     & & & instruction with lifting the said object. \\
    \midrule

    Put X near Y     & $3 \times 3$ & 0 & The agent is spawned randomly. \\
    (X = Set $A$ or $B$, & & & Room has $1$ object from Set $A$ (or $B$) and $1$ from \\
     Y = Set $C \cup D$) & & & $C \cup D$ and instructed to put the object from Set $A$ (or $B$)  \\
         & & & near the other. The purpose of these training tasks is to learn to \\ 
      & & & move one object near another before putting it on it. \\
    \midrule

    Put X on Y       & $3 \times 3$ & 0 & The agent is spawned randomly. \\
    (X = Set $A$ or $B$, & & & Room has $1$ object from Set $A$ (or $B$) and $1$ from \\
     Y = Set $C \cup D$) & & & $C \cup D$ and instructed to put the object from Set $A$ (or $B$)  \\
          & & & on top of the other. The purpose of these training tasks is to learn to \\ 
      & & & move one object and place it on top of another. \\
    \midrule

    Put X on Y       & $4 \times 4$ & 4 & The agent is spawned randomly. \\
    (X = $A$, Y = $D$ & & & Room has $3$ objects from Set $A$ (or $B$) and $3$ from \\
     or & & &  Set $D$ (or $C$) and instructed to put the object from Set $A$ (or $B$)  \\
     X = $B$, Y = $C$) & & & on top of the other. This is the training task most similar to the \\ 
                        & & & test task and requires mastering all the other ones. \\
    \midrule

    Put X on Y \textbf{(test)}  & $4 \times 4$ & 4 & The agent is spawned randomly. \\
    (X = $A$, Y = $C$ & & & Room has $3$ objects from Set $A$ (or $B$) and $3$ from \\
     or             & & &  Set $C$ (or $D$) and instructed to put the object from Set $A$ (or $B$)  \\
     X = $B$, Y = $D$) & & & on top of the other. This is the test task. \\
    \bottomrule
    \end{tabular}
  \end{center}
\end{table}

\begin{table}
  \begin{center}
    \caption{\label{tab:playroom-action-space} 3d Unity Environment action space.}
    \begin{tabular}{ll}
    \toprule
    \textbf{General body movement} & \textbf{Fine grain movement} \\
    \midrule
    NOOP & MOVE\_RIGHT\_SLIGHTLY \\
    MOVE\_FORWARD\_FULL & MOVE\_LEFT\_SLIGHTLY \\
    MOVE\_BACKWARD\_FULL & LOOK\_RIGHT\_MID \\ 
    MOVE\_RIGHT\_FULL & LOOK\_LEFT\_MID \\
    MOVE\_LEFT\_FULL & LOOK\_DOWN\_MID \\
    LOOK\_RIGHT\_FULL & LOOK\_UP\_MID \\
    LOOK\_LEFT\_FULL & LOOK\_RIGHT\_SLIGHTLY \\
    LOOK\_DOWN\_FULL & LOOK\_LEFT\_SLIGHTLY \\
    LOOK\_UP\_FULL &  \\
    \midrule
    \textbf{Fine grained movement with grip} & \textbf{General body movement with grip} \\
    \midrule
    GRAB + MOVE\_RIGHT\_MID & GRAB \\
    GRAB + MOVE\_LEFT\_MID & GRAB + MOVE\_FORWARD\_FULL \\
    GRAB + LOOK\_RIGHT\_MID & GRAB + MOVE\_BACKWARD\_FULL \\
    GRAB + LOOK\_LEFT\_MID & GRAB + MOVE\_RIGHT\_FULL \\
    GRAB + LOOK\_DOWN\_MID & GRAB + MOVE\_LEFT\_FULL \\
    GRAB + LOOK\_UP\_MID & GRAB + LOOK\_RIGHT\_FULL \\
    GRAB + LOOK\_RIGHT\_SLIGHTLY & GRAB + LOOK\_LEFT\_FULL \\
    GRAB + LOOK\_LEFT\_SLIGHTLY & GRAB + LOOK\_DOWN\_FULL \\
    GRAB + PULL\_CLOSER\_MID & GRAB + LOOK\_UP\_FULL \\
    GRAB + PUSH\_AWAY\_MID & \\
    \midrule
    \textbf{Object manipulation} & \\
    \midrule
    GRAB + SPIN\_RIGHT  & \\
    GRAB + SPIN\_LEFT & \\
    GRAB + SPIN\_UP & \\
    GRAB + SPIN\_DOWN & \\
    GRAB + SPIN\_FORWARD & \\
    GRAB + SPIN\_BACKWARD & \\
    GRAB + PULL\_CLOSER\_FULL & \\
    GRAB + PUSH\_AWAY\_FULL & \\
    PULL\_CLOSER\_MID & \\
    PUSH\_AWAY\_MID & \\
    
    \bottomrule
    \end{tabular}
  \end{center}
\end{table}

\end{document}